\documentclass[letterpaper]{article} 
\usepackage{aaai24}  
\usepackage{times}  
\usepackage{helvet}  
\usepackage{courier}  
\usepackage[hyphens]{url}  
\usepackage{graphicx} 
\usepackage{natbib}  
\usepackage{caption} 
\usepackage{algorithm}
\usepackage{algorithmic}
\usepackage{xcolor}
\usepackage{soul}
\usepackage{amsfonts}
\usepackage{amsmath}
\usepackage{amsthm}
\usepackage[final]{pdfpages}

\newcommand{\wbigcup}{\mathop{\bigcup}\displaylimits}

\usepackage{newfloat}
\usepackage{listings}
\DeclareCaptionStyle{ruled}{labelfont=normalfont,labelsep=colon,strut=off} 
\floatstyle{ruled}
\newfloat{listing}{tb}{lst}{}
\floatname{listing}{Listing}
\title{Cultivating Archipelago of Forests: \\Evolving Robust Decision Trees through Island Coevolution}
\author{
  Adam {\.Z}ychowski\textsuperscript{\rm 1},
  Andrew Perrault\textsuperscript{\rm 2},
  Jacek Ma{\'n}dziuk\textsuperscript{\rm 1, 3, 4}
}
\affiliations{
  \textsuperscript{\rm 1}Faculty of Mathematics and Information Science, Warsaw University of Technology \\
  \textsuperscript{\rm 2}Department of Computer Science and Engineering, The Ohio State University\\
  \textsuperscript{\rm 3}Faculty of Computer Science, AGH University of Krakow \\
  \textsuperscript{\rm 4}Center of Excellence in Artificial Intelligence, AGH University of Krakow \\
  adam.zychowski@pw.edu.pl, perrault.17@osu.edu, jacek.mandziuk@pw.edu.pl
}

\begin{document}

\maketitle

\begin{abstract}
Decision trees are widely used in machine learning due to their simplicity and interpretability, but they often lack robustness to adversarial attacks and data perturbations. The paper proposes a novel island-based coevolutionary algorithm (ICoEvoRDF) for constructing robust decision tree ensembles. The algorithm operates on multiple islands, each containing populations of decision trees and adversarial perturbations. The populations on each island evolve independently, with periodic migration of top-performing decision trees between islands. This approach fosters diversity and enhances the exploration of the solution space, leading to more robust and accurate decision tree ensembles. ICoEvoRDF utilizes a popular game theory concept of mixed Nash equilibrium for ensemble weighting, which further leads to improvement in results.
ICoEvoRDF is evaluated on 20 benchmark datasets, demonstrating its superior performance compared to state-of-the-art methods in optimizing both adversarial accuracy and minimax regret. The flexibility of ICoEvoRDF allows for the integration of decision trees from various existing methods, providing a unified framework for combining diverse solutions. Our approach offers a promising direction for developing robust and interpretable machine learning models.
\end{abstract}

%

\section{Introduction}

Decision trees (DTs) are a popular tool in the field of machine learning due to their inherent simplicity, interpretability, and efficiency. Their capacity to model complex decision boundaries in an understandable manner has made them a popular choice for both academic research and practical applications~\cite{rokach2005decision}. However, single DTs often struggle with issues such as overfitting, lack of robustness to noise, and limited generalization capabilities. To mitigate these problems, ensembles of DTs~\cite{banfield2006comparison}, such as random forests, have been developed that aggregate multiple trees to improve overall performance.

Despite these advancements in constructing robust and accurate DT ensembles, several challenges remain. Traditional methods for creating robust decision trees (RDTs) and robust decision forests (RDFs) often rely on only one particular robustness metric, limiting the practical applicability of these methods in real-world scenarios. Balancing multiple objectives, such as accuracy, robustness, and computational efficiency is often not adequately addressed. Existing methods also often struggle to maintain diversity within the ensemble, leading to reduced effectiveness.

We aim to develop better methods for building and combining decision tree ensembles (RDFs). Evolutionary algorithms~\cite{michalewicz2013genetic}, which mimic natural selection to solve complex problems, have demonstrated significant potential in optimizing complex, high-dimensional search spaces~\cite{yu2010introduction}, making them a suitable candidate for evolving decision tree ensembles. However, existing evolutionary approaches operate only with a single population and produce single decision trees~\cite{zychowski2024CoEvoRDT}.

To address these limitations, we propose a novel island-based coevolutionary approach named ICoEvoRDF (Island-based CoEvolutionary Robust Decision Forests). The approach is inspired by island models~\cite{tanese1989distributed,skolicki2004improving} in evolutionary computation. Island models subdivide a population into several isolated subpopulations, each evolving independently. Occasionally, individuals migrate between islands, introducing beneficial traits and enhancing overall diversity~\cite{bull2007learning}. This approach has been successful in various optimization problems, often performing better than traditional single-population methods~\cite{skolicki2004improving,skolicki2005analysis,luong2010gpu}.

In the context of DT ensembles, the use of island models can be particularly advantageous. By evolving multiple populations of RDTs in parallel, each with different initial conditions and evolutionary trajectories, we can generate a diverse set of trees that, when combined, form a more robust and versatile forest. Moreover, this approach may also be advantageous for creating single RDTs, as previous studies~\cite{zychowski2024CoEvoRDT} have demonstrated significant benefits from repeated runs of evolutionary methods.

Additionally, we employ the game-theoretic concept of Mixed Nash Equilibrium (MNE) to effectively weight the decision trees in the forest. Unlike uniform weighting methods, MNE optimizes tree contributions by reflecting the full set of perturbations encountered in adversarial settings. The island-based coevolutionary framework naturally generates diverse input data perturbations, which are crucial for capturing the range of scenarios represented by the MNE. Integrating MNE into our coevolutionary framework enhances robustness and performance, demonstrating the power of combining game theory with coevolutionary methods.

\subsection{Contributions}
The main contributions of the work are:
\begin{itemize}
\item a novel island-based coevolutionary approach (ICoEvoRDF) for creating a mixture of Robust Decision Trees (RDTs), which utilizes a new game-theoretic approach based on Mixed Nash Equilibrium~\cite{fudenberg1993learning} for weighting decision trees within a forest. The method is 
adaptable 
to optimizing various target metrics, such as adversarial accuracy and minimax regret. Furthermore, it is able to  
construct individual RDTs, that are superior to those constructed by  evolutionary-based approaches with a single population;
\item potential for ICoEvoRDF to improve results of other state-of-the-art (SOTA) methods by including their resulting decision trees in the initial population of one or more islands, thus allowing for the combination of multiple solutions into a single, synergetic framework;
\item comprehensive analysis of how specific components of the proposed solution (e.g., number of islands, Nash-based voting, various initialization methods, population diversity) influence its performance and outcomes,
\item experimental validation showing that ICoEvoRDF significantly outperforms existing methods for ensemble of trees as well as single DTs, with improvements of up to 0.010 in adversarial accuracy and 0.064 in minimax regret across 20 benchmark datasets.
\end{itemize}

\section{Problem definition}

Let $X \subset \mathbb{R}^d$ be a $d$-dimensional feature space (inputs) and $Y$ be the set of possible classes (outputs). A standard classification task involves finding a function (model) $h: X \rightarrow Y$, where $h(x_i) = y_i$ and $y_i$ is the true class of $x_i$. The classification performance of $h$ can be measured by its accuracy, defined as:
\begin{align}\mathrm{acc}(h) = \frac{1}{|X|}\sum_{x_i \in X} I[h(x_i) = y_i],\end{align}
where $I[h(x_i) = y_i]$ returns 1 if $h$ predicts the true class of $x_i$ and 0 otherwise.

Define $\mathcal{N}_\varepsilon(x) = \{ z : ||z - x||_\infty \leq \varepsilon \}$ as the set of points within an $L_\infty$ ($L$-infinity norm) ball centered at $x$ with radius $\varepsilon$. The \textit{adversarial accuracy} of a model $h$ is the lowest accuracy over all possible perturbation sets constrained by $\mathcal{N}_\varepsilon$, formally:
\begin{align}\mathrm{acc}_{\mathrm{adv}}(h, \varepsilon) = \frac{1}{|X|}\sum_{x_i \in X} \min_{z_i \in \mathcal{N}_\varepsilon(x_i)} I[h(z_i) = y_i].\end{align}

If $\varepsilon$ is large enough, it may not be possible to train $M$ effectively, and the adversarial accuracy would be low. However, low adversarial accuracy does not indicate whether the reason is poor model robustness or impossibility of creating such a model due to the high value of $\varepsilon$.

The \textit{max regret}~\cite{savage1951theory} of a model $h$ is the maximum regret among all possible perturbations $z \in \mathcal{N}_\varepsilon$. Regret is defined as the difference between the best possible accuracy on a particular perturbation and the accuracy that $h$ achieves:
\begin{align}\mathrm{regret}(h, \{z_i\}) = \max_{h'} \mathrm{acc}(h', \{z_i\}) - \mathrm{acc}(h, \{z_i\}),\end{align}
where $\mathrm{acc}(h, \{z_i\})$ is the accuracy achieved by model $h$ when the feature set $\{x_i\}$ is replaced with $\{z_i\}$. Max regret can be expressed as:
\begin{align}\mathrm{mr}(h) = \max_{z_i \in \mathcal{N}_\varepsilon(x_i)} \mathrm{regret}(h, \{z_i\}).\end{align}
Max regret is defined as the maximum difference between the result of a given model and the result of the optimal model for any input data perturbation within a given range. Minimizing max regret is known as the minimax regret decision criterion.
Minimax regret is a more comprehensive metric than adversarial accuracy because it takes into account the performance of a model on both clean (non-perturbed) and adversarial (perturbed) data, so it is less sensitive to the choice of $\varepsilon$.

The problem addressed in this paper is finding a decision forest (ensemble of decision trees) trained on $X$ that for a given $\varepsilon$ optimizes (maximizes for adversarial accuracy or minimizes for max regret) a given robustness metric. The final prediction from decision trees ensemble are obtained by the weighted voting. 

We use \textit{Mixed Nash Equilibrium} (MNE) for computing the weights of trees in an ensemble. An MNE $(\mathcal{T}, \mathcal{P})$ comprises a pair of mixed strategies: $\mathcal{T} = \{(T_1, p_{T_1}), \ldots, (T_n, p_{T_n})\}$, and $\mathcal{P} = \{(P_1, p_{P_1}), \ldots, (P_m, p_{P_m})\}$. In our case, those are a set of DTs ($\mathcal{T}$) and perturbations ($\mathcal{P}$), respectively, with associated probabilities.  $(\mathcal{T},\mathcal{P})$ fulfills the following conditions: $\forall_{\mathcal{T'} \neq \mathcal{T}}, \mathbb{E}_{\mathcal{T}, \mathcal{P}} [\xi(\mathcal{T'},\mathcal{P})] \preceq \mathbb{E}_{\mathcal{T}, \mathcal{P}} [\xi(\mathcal{T},\mathcal{P})]$ and $\forall_{\mathcal{P'} \neq \mathcal{P}}, \mathbb{E}_{\mathcal{T}, \mathcal{P}}[-\xi(\mathcal{T},\mathcal{P'})] \preceq \mathbb{E}_{\mathcal{T}, \mathcal{P}}[-\xi(\mathcal{T},\mathcal{P})]$, where $\xi(\mathcal{T},\mathcal{P})$ is some performance metric calculated for a ``mixed'' decision tree $\mathcal{T}$ and ``mixed'' perturbation $\mathcal{P}$ (accuracy for adversarial accuracy or regret for max regret).


Note that the game is zero-sum because in robust optimization the adversary aims to minimize the DT payoff.
Thus, the minimax theorem holds~\cite{v1928theorie}, guaranteeing that the MNE maximizes the robustness metric, i.e., $\max_\mathcal{T'} \min_\mathcal{P'} \xi(\mathcal{T'},\mathcal{P'}) = \mathbb{E}_{\mathcal{T}, \mathcal{P}}[\xi(\mathcal{T},\mathcal{P})]$ if $(\mathcal{T}, \mathcal{P})$ is an MNE. In addition, for zero-sum games, MNEs can be computed in time polynomial in the strategy set size.

\section{Related work}
\label{sec:related-work}

The problem of finding RDTs is well-established in the literature, and several methods of solving it have been proposed. RIGDT-h~\cite{chen2019robust} constructs robust decision trees based on the concept of adversarial Gini impurity~\cite{breiman2017classification}, which is a modification of the classical Gini impurity adapted for perturbed input data. 
This method was further improved in the GROOT algorithm~\cite{vos2021efficient}, which employs a greedy recursive splitting strategy similar to traditional decision trees and evaluates splits using the adversarial Gini impurity.
Another algorithm, Fast Provably Robust Decision Tree (FPRDT)~\cite{guo2022fast}, builds robust decision trees by directly minimizing the adversarial 0/1 loss through a greedy recursive approach, efficiently evaluating potential splits using sorted threshold lists. The algorithm ensures optimal splits only when beneficial, maintaining both robustness and efficiency.

Since finding RDTs is an optimization problem, some evolutionary-based methods were also proposed. \citet{ranzato2021genetic} introduced a genetic adversarial training algorithm (Meta-Silvae) to optimize DT stability.
Coevolutionary Algorithm for Robust Decision Trees (CoEvoRDT)~\cite{zychowski2024CoEvoRDT}  constructs robust decision trees by maintaining two populations: one of decision trees and another of data perturbations, allowing the trees to adapt and learn from the perturbations. The populations alternately evolve using crossover, mutation, and selection operators until a stopping condition is met. The method emphasizes robustness by incorporating a game-theoretic approach to build a Hall of Fame (an archive that preserves the best solutions found during the algorithm's run) with a Mixed Nash Equilibrium, enhancing decision tree robustness and population diversity. CoEvoRDT demonstrates strong performance in minimax regret and adversarial accuracy, making it one of the state-of-the-art algorithms.

Tree ensembles were also explored in the context of finding robust classifiers as a natural subsequent step. 
\citet{kantchelian2016evasion} introduced a method to iteratively and recursively train robust decision trees using the $L_0$ metric based on the original and adversarial examples. \citet{vos2021efficient} proposed robust random forests based on the original idea of random forests~\cite{breiman2001random}. \citet{andriushchenko2019provably} use a robust boosting algorithm based on adversarial exponential loss.
Provably Robust AdaBoost (PRAdaBoost)~\cite{guo2022fast} method is built upon the AdaBoost algorithm~\cite{freund1996experiments}, a well-established method in machine learning that combines multiple "weak" learners to create a "strong" classifier. PRAdaBoost utilizes the FPRDT algorithm mentioned above as the base learner.

To the best of our knowledge, so far no other multi-population approaches have been proposed for optimizing RDTs or RDFs, and our method is the first successful attempt in this area.

\section{ICoEvoRDF algorithm}

In this study, we introduce the Island-based CoEvolutionary algorithm for constructing Robust Decision Forests (ICoEvoRDF). This approach draws inspiration from the biological concept of speciation, where distinct populations evolve in isolated environments with limited gene flow.

ICoEvoRDF operates on a fixed set of islands, denoted as $\mathcal{I}$, whose cardinality $|\mathcal{I}|$ is a user-defined parameter. Each island $I \in \mathcal{I}$ maintains two coevolving populations: a decision tree population $I^T$ and a perturbation population $I^P$, analogously to the CoEvoRDT algorithm detailed in the previous section. A pseudocode of the proposed algorithm is presented in the supplementary material.

\vspace{0.2em}\noindent\textbf{Islands initialization.}
We initialize each island independently. For the baseline implementation, the decision tree population is initialized with small random trees (with a maximum depth of 3). The potential advantages of alternative initialization methods, such as utilizing the results of other algorithms are discussed in Section~\ref{sec:results}.

The perturbation population is sampled uniformly from the space of all possible perturbations of the input data within a given radius $\epsilon$. For input data selection, we propose to assign a unique training set to each island. It mirrors the approach of classic random forests and is achieved by sampling with replacement from $X$ until each island's training set contains $|X|$ instances. This approach permits the repetition of certain training instances across multiple islands.

\vspace{0.2em}\noindent\textbf{Islands evolution.}
Each island evolves independently for a predetermined number of generations, denoted by $n_g$. This process entails coevolution between the DT and perturbation populations, similarly to the CoEvoRDT algorithm. The DT and perturbation populations evolve alternately for $n_g$ generations, each involving standard evolutionary operations: crossover, mutation, evaluation, and selection. 
The mutation operator randomly performs one of the three following actions: (i) replacing a subtree with a randomly generated one, (ii) changing the information in a randomly selected node (e.g., a new splitting value or operator), or (iii) pruning a randomly selected subtree. The crossover operator randomly selects one node in each of 
two individuals and exchanges the corresponding subtrees, generating two offspring added to the population.
In the perturbation population, each pair (input instance, attribute) has a 
probability 0.5 of being perturbed, with a new feasible value assigned (according to the epsilon constraint). The crossover 
randomly mixes input instances from both individuals.

Evaluation is conducted against individuals from the opposing population (DTs are evaluated against perturbations and vice versa). Furthermore, a Hall of Fame (HoF) mechanism inspired by Nash equilibrium theory has been incorporated into ICoEvoRDF, which proved to be beneficial in CoEvoRDT method. HoF maintains a mixture of individuals from both populations which are used during the evaluation stage. The final stage involves binary tournament selection, which selects individuals for the next generation. In this process, individuals are randomly paired, and the one with higher fitness has a greater probability of being chosen. The size of the population is fixed, i.e., in each generation the same number of individuals are promoted.

\vspace{0.2em}\noindent\textbf{Migration.}
Migration between islands is an important component of island-based evolutionary algorithms, differentiating them from $|\mathcal{I}|$ independent runs of a single island. If subpopulations evolve in complete isolation (without migration), they might converge prematurely to suboptimal solutions due to the lack of new genetic material~\cite{gong2015distributed}. Migration introduces new genetic traits, reducing the risk of premature convergence and potentially leading to superior solutions through the sharing of beneficial traits.

\begin{figure*}[ht]
\centering
\includegraphics[width=0.85\textwidth]
{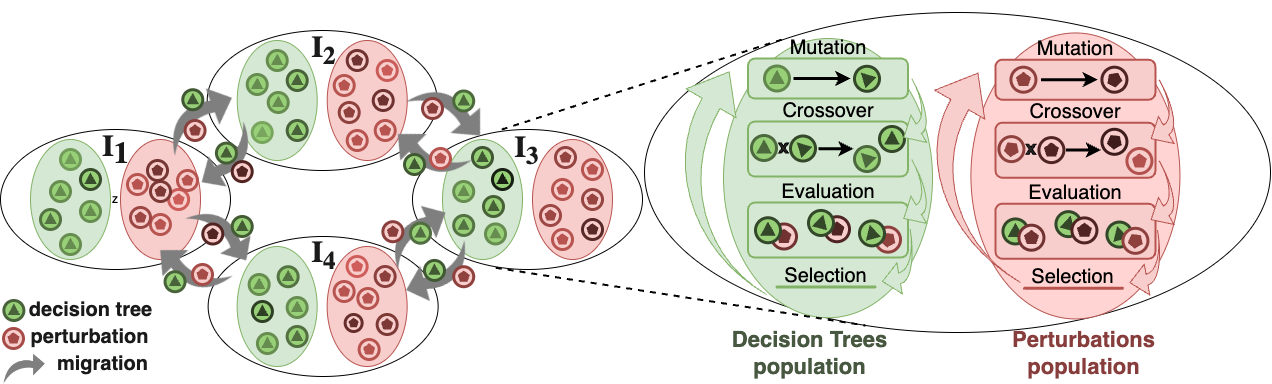}
\caption{The ICoEvoRDF scheme. The illustration shows 4 islands with ring migration topology. Each island contains two populations: DTs and perturbations which are developed alternately using evolutionary operators.}
\label{fig:ICoEvoRDF-scheme}
\end{figure*}

For each island $I$, we define a set of neighboring islands $\eta(I)$ to which $I$ transmits information about its discovered solutions. The connections between islands and their neighbors constitute a graph referred to as the island topology. For instance, a ring topology forms a bidirectional cycle where each island possesses exactly two neighbors: $\forall_{I \in \mathcal{I}} |\eta(I)|=2$. Figure~\ref{fig:ICoEvoRDF-scheme} provides an overview of the ICoEvoRDF algorithm and illustrates an example of a ring topology.

After every $n_g$ generations of both the DT and perturbation populations, the top $k_{\mathrm{top}}$ fittest (i.e., highest fitness) decision trees from each neighboring island $I_n \in \eta(I)$ are migrated (copied) into the recipient island's DT population:
\begin{align}I^T = I^T \cup \wbigcup_{I_n \in \eta(I)} V^{k_{\mathrm{top}}}_\xi(I_n^T),    
\end{align} where $V^{k_{\mathrm{top}}}_\xi(I_n^T)$ is the $k_{\mathrm{top}}$ fittest decision trees from the DT population of the neighbor island $I_n^T$ according to the optimized metric $\xi$.
The $k_{\mathrm{top}}$ fittest perturbations are migrated in the same way.

\vspace{0.2em}\noindent\textbf{Stop condition.}
This iterative process persists until at least one of the following two conditions is fulfilled: either a total of $l_g$ generations have elapsed, or the fittest DT across all islands has not improved within the last $l_c$ generations.

Upon termination, either a single, globally fittest DT is selected across all islands, or an ensemble of DTs from all islands are aggregated to construct a decision forest (DF).

\vspace{0.2em}\noindent\textbf{Decision forest composition.}
For creation of a DT ensemble we perform a weighted voting among the fittest DT representatives from all islands.The simplest and most common approach is to assign equal contribution to each island representative. However, it may not be optimal, as the efficacy of the generated trees may vary, rendering equal contributions undesirable. Leveraging the fact that each island also generates a population of perturbations, we propose \textit{Nash-based voting} (NV). This approach frames the scenario as a two-player game between DTs (represented by island representatives) and perturbations. The DT player, $P_{DT}$, chooses their strategy from a strategy set $\Pi_{P_{DT}} = \{V^1_\xi(I^T)\}_{I \in \mathcal{I}}$, while the perturbation player, $P_P$, chooses their strategy from a strategy set $\Pi_{P_{P}} = \bigcup_{I \in \mathcal{I}} I^P$. Then, a mixed Nash equilibrium is computed for this game. We use the Lemke-Howson algorithm~\cite{lemke1964equilibrium}, which has worst-case exponential time complexity, but is fast in practice (and has average linear time for random games~\cite{codenotti2008experimental}).
In the proposed Nash-based voting (\textit{NV}) probabilities from the mixed equilibrium DT strategy $\mathcal{T}$ are directly used for voting weights.

We also explored an alternative approach to extracting from each island for the final ensemble more DTs than only the fittest one. 
These experiments resulted in only marginal improvement (less than 0.1\%), coupled with increased computational overhead and complexity of the final model.

\section{Experimental setup}
\label{sec:experimental-setup}
\textbf{Tested benchmarks.} 
The proposed method was evaluated on 20 widely used classification benchmark problems with varying characteristics, including the number of instances (from 351 to 70000), features (4 -- 3072), and perturbation coefficients $\varepsilon$ (0.01 -- 0.3). All selected datasets have been utilized in prior studies referenced in the Related Work section and are publicly available at https://www.openml.org. 
The main parameters of the datasets 
(perturbation coefficient, number of instances, features, and classes) are presented in supplementary material.

\vspace{0.2em}\noindent\textbf{Parameterization.} 
For the experiments conducted in this study, we employed the same set of parameters across all islands and their values followed recommendation from CoEvoRDT authors. Namely, decision tree population size $N_T = 200$, perturbation population size $N_P = 500$, number of consecutive generations for each population $n_p = 20$, number of best individuals from the DT population involved in the perturbations evaluation $N_\mathrm{\mathrm{top}} = 20$, crossover probability $p_c = 0.8$, mutation probability $p_m = 0.5$, selection pressure $p_s = 0.9$, elite size $e = 2$, HoF size $N_\mathrm{HoF} = 200$.
Generations without improvement limit $l_c$ was set to 100 and generation limit $l_g = 1000$.

Preliminary evaluations were conducted to assess various migration topologies, such as ring, star, and circle configurations. The ring topology (depicted in Figure~\ref{fig:ICoEvoRDF-scheme}) yielded the most favorable results and was thus used for all experiments presented in the subsequent section. A more comprehensive discussion and detailed results for different topologies can be found in the supplementary material. The number of generations between migrations, $n_g$, was set to 40.

The number of islands, $|\mathcal{I}|$, was fixed at 10 to maintain comparable computational time to the state-of-the-art (SOTA) method, PRAdaBoost. However, our experiments suggest that further increases in the DT population size ($N_T$) or the number of islands $|\mathcal{I}|$ may lead to enhanced performance. A detailed analysis of the relationship between population size/number of islands, computational time, and robustness can be found in the supplementary material. 
This is also related to interpretability which is also important aspect in the context of DTs. While adding more DTs to the forest can further (slightly) enhance robustness, it also deteriorates the model's overall interpretability.

All results for nondeterministic methods presented in the next section represent the average of 20 independent runs. We refer the reader to the supplementary material for detailed results about their variations (standard deviations). Statistical significance was checked according to the paired $t$-test with $p$-value $\leq 0.05$. All tests were executed on an Intel Xeon Silver 4116 processor operating at 2.10GHz. The CoEvoRDT source code is available at github.com/zychowskia/ICoEvoRDF. We employed the Nashpy Python library~\cite{nashpy} implementation of the Lemke-Howson algorithm for computing the Mixed Nash Equilibrium. The CART method~\cite{breiman2017classification} was used to determine the reference tree for minimax regret (the tree with the highest accuracy for a given perturbation). The final results for minimax regret were computed based on $10^5$ randomly generated perturbations (see supplementary material for reasoning) -- the same for all tested methods. 
The 
adversarial accuracy results were calculated 
using the exact method based on Mixed Integer Linear Programming~\cite{kantchelian2016evasion}, implemented by~\cite{chen2019robust}.

\section{Results and discussion}
\label{sec:results}

\vspace{0.2em}\noindent\textbf{Trees ensemble models.}
Tables~\ref{tab:DF-accuracy} and~\ref{tab:DF-regret} present the results for ensemble methods. GROOT, FPRDT, and CoEvoRDT construct decision forests based on their respective underlying DT creation algorithms, analogously to the standard random forest algorithm.

PRAdaBoost and CoEvoRDT boosting are boosting methods that combine multiple weak learners (FPRDT and CoEvoRDT, respectively) to create a stronger model. Each DT is trained sequentially, with an emphasis on rectifying errors made by previous trees. Misclassified instances are assigned greater weights to enhance overall accuracy. The final model is a weighted sum of all DTs.

In terms of adversarial accuracy, the proposed ICoEvoRDF algorithm achieved superior results on 17 benchmarks, for 9 of them with statistical significance. The most competitive non-evolutionary method, PRAdaBoost, yielded the highest score on two datasets. With respect to the max regret metric, the proposed approach outperformed all other methods across all benchmark datasets, in each case with statistical significance).

We also explored the combination of the strongest non-evolutionary method, FPRDT, with the proposed coevolutionary islanding approach. This was accomplished by initializing one island with the FPRDT algorithm (denoted by the suffix ``+ FPRDT"), while all other islands were initialized with random DTs. For some datasets (particularly those where FPRDT performed well, such as the diabetes dataset) they further improved the results.

A comprehensive analysis of standard deviations (std dev) for each method is available in the supplementary material. Notably, among the top-performing methods, PRAdaBoost exhibits an average std dev of 0.0085, while the ICoEvoRDF family of methods demonstrates a std dev of 0.0100.
\begin{table*}[ht]
\setlength{\tabcolsep}{3pt}
\renewcommand{\arraystretch}{1.01}
\resizebox{0.95\textwidth}{!}{
\begin{tabular}{|l|c|c|c|c|c|c|c|c|c|c|c|}
\hline
\textbf{dataset} & \begin{tabular}[c]{@{}c@{}}Random \\ forests\end{tabular} & \begin{tabular}[c]{@{}c@{}}\small GROOT \\ \small forests\end{tabular} & \begin{tabular}[c]{@{}c@{}}\small FPRDT \\ \small forest\end{tabular} & \begin{tabular}[c]{@{}c@{}}\small CoEvoRDT \\ \small forest\end{tabular} & \small PRAdaBoost & \begin{tabular}[c]{@{}c@{}}\small CoEvoRDT \\ \small boosting\end{tabular} & \small ICoEvoRDF$\phantom{}^{EV}_{SI}$ & \small ICoEvoRDF$\phantom{}_{SI}$ & \small ICoEvoRDF$\phantom{}^{EV}$ & \small ICoEvoRDF & \begin{tabular}[c]{@{}c@{}}\small ICoEvoRDF\\ \small + FPRDT\end{tabular} \\ \hline
ionos & 0.112 & 0.787 & 0.791 & 0.793 & 0.796 & 0.798 & 0.797 & 0.796 & 0.796 & 0.799 & \textbf{0.801} \\ 
breast & 0.217 & 0.884 & 0.873 & 0.885 & 0.879 & 0.899 & 0.891 & 0.894 & 0.896 & \textbf{0.900} & \textbf{0.900} \\ 
diabetes & 0.452 & 0.648 & 0.649 & 0.621 & \textbf{0.654} & 0.644 & 0.625 & 0.636 & 0.646 & 0.647 & 0.651 \\ 
bank & 0.509 & 0.641 & 0.658 & 0.661 & 0.668 & 0.669 & 0.667 & 0.670 & 0.664 & \textbf{0.673} & 0.672 \\
Japan3v4 & 0.519 & 0.658 & 0.669 & 0.679 & 0.682 & 0.684 & 0.684 & 0.688 & 0.684 & 0.688 & \textbf{0.690} \\ 
spam & 0.000 & 0.750 & 0.749 & 0.751 & 0.754 & 0.763 & 0.756 & 0.756 & 0.762 & \textbf{0.766} & \textbf{0.766} \\ 
GesDvP & 0.189 & 0.731 & 0.725 & 0.740 & 0.732 & 0.753 & 0.745 & 0.745 & 0.749 & 0.752 & \textbf{0.754} \\ 
har1v2 & 0.233 & 0.792 & 0.828 & 0.844 & \textbf{0.860} & 0.851 & 0.855 & 0.858 & 0.847 & 0.854 & \textbf{0.860} \\ 
wine & 0.091 & 0.633 & 0.681 & 0.688 & 0.690 & \textbf{0.708} & 0.691 & 0.691 & 0.707 & \textbf{0.708} & \textbf{0.708} \\
collision-det & 0.325 & 0.726 & 0.791 & 0.804 & 0.800 & 0.820 & 0.810 & 0.812 & 0.815 & \textbf{0.822} & \textbf{0.822} \\ 
mnist-1-5 & 0.000 & 0.925 & 0.964 & 0.964 & 0.969 & 0.975 & 0.969 & 0.972 & 0.968 & \textbf{0.976} & \textbf{0.976} \\ 
mnist-2-6 & 0.000 & 0.823 & 0.919 & 0.917 & 0.924 & 0.925 & 0.923 & 0.925 & 0.922 & \textbf{0.926} & \textbf{0.926} \\ 
mnist & 0.000 & 0.632 & 0.750 & 0.747 & 0.761 & 0.763 & 0.755 & 0.759 & 0.759 & \textbf{0.764} & \textbf{0.764} \\ 
F-mnist2v5 & 0.456 & 0.979 & 0.974 & 0.982 & 0.982 & 0.993 & 0.990 & 0.994 & 0.987 & 0.995 & \textbf{0.996} \\ 
F-mnist3v4 & 0.044 & 0.839 & 0.861 & 0.869 & 0.867 & 0.879 & 0.877 & 0.877 & 0.877 & \textbf{0.884} & \textbf{0.884} \\ 
F-mnist7v9 & 0.136 & 0.836 & 0.875 & 0.868 & 0.879 & 0.877 & 0.877 & 0.880 & 0.873 & \textbf{0.881} & 0.880 \\ 
F-mnist & 0.024 & 0.241 & 0.537 & 0.545 & 0.546 & 0.559 & 0.552 & 0.553 & 0.554 & 0.560 & \textbf{0.561} \\ 
cifar10:0v5 & 0.302 & 0.526 & 0.683 & 0.690 & 0.691 & 0.699 & 0.694 & 0.696 & 0.697 & 0.702 & \textbf{0.703} \\
cifar10:0v6 & 0.368 & 0.560 & 0.688 & 0.696 & 0.696 & 0.703 & 0.701 & 0.701 & 0.701 & 0.704 & \textbf{0.705} \\ 
cifar10:4v8 & 0.296 & 0.498 & 0.665 & 0.665 & 0.671 & 0.671 & 0.674 & 0.674 & 0.673 & \textbf{0.675} & \textbf{0.675} \\ \hline
\textbf{AVERAGE} & 0.214 & 0.705 & 0.767 & 0.771 & 0.775 & 0.782 & 0.777 & 0.779 & 0.779 & 0.784 & \textbf{0.785} \\ \hline
\end{tabular}
}
\caption{
Averaged adversarial accuracies for ensemble forests methods. The best results are bolded.}
\label{tab:DF-accuracy}
\end{table*}
\begin{table*}[ht]
\setlength{\tabcolsep}{3pt}
\renewcommand{\arraystretch}{1.01}
\resizebox{0.95\textwidth}{!}{
\begin{tabular}{|l|c|c|c|c|c|c|c|c|c|c|c|}
\hline
\textbf{dataset} & \begin{tabular}[c]{@{}c@{}}Random \\ forests\end{tabular} & \begin{tabular}[c]{@{}c@{}}\small GROOT \\ \small forests\end{tabular} & \begin{tabular}[c]{@{}c@{}}\small FPRDT \\ \small forest\end{tabular} & \begin{tabular}[c]{@{}c@{}}\small CoEvoRDT \\ \small forest\end{tabular} & \small PRAdaBoost & \begin{tabular}[c]{@{}c@{}}\small CoEvoRDT \\ \small boosting\end{tabular} & \small ICoEvoRDF$\phantom{}^{EV}_{SI}$ & \small ICoEvoRDF$\phantom{}_{SI}$ & \small ICoEvoRDF$\phantom{}^{EV}$ & \small ICoEvoRDF & \begin{tabular}[c]{@{}c@{}}\small ICoEvoRDF\\ \small + FPRDT\end{tabular} \\ \hline
ionos & 0.094 & 0.088 & 0.061 & 0.052 & 0.060 & 0.045 & 0.046 & 0.046 & 0.048 & \textbf{0.044} & \textbf{0.044} \\
breast & 0.103 & 0.097 & 0.055 & 0.047 & 0.055 & 0.035 & 0.047 & 0.044 & 0.039 & \textbf{0.034} & \textbf{0.034} \\
diabetes & 0.202 & 0.194 & 0.111 & 0.092 & 0.113 & 0.027 & 0.076 & 0.045 & 0.029 & \textbf{0.026} & \textbf{0.026} \\
bank & 0.186 & 0.177 & 0.086 & 0.075 & 0.086 & 0.050 & 0.051 & 0.050 & 0.061 & \textbf{0.046} & \textbf{0.046} \\
Japan3v4 & 0.107 & 0.106 & 0.063 & 0.060 & 0.062 & 0.027 & 0.028 & 0.025 & 0.032 & \textbf{0.025} & \textbf{0.025} \\
spam & 0.097 & 0.095 & 0.071 & 0.067 & 0.071 & 0.046 & 0.063 & 0.061 & 0.054 & \textbf{0.044} & \textbf{0.044} \\
GesDvP & 0.152 & 0.143 & 0.127 & 0.111 & 0.122 & 0.077 & 0.110 & 0.108 & 0.090 & \textbf{0.073} & \textbf{0.073} \\
har1v2 & 0.105 & 0.100 & 0.066 & 0.063 & 0.061 & 0.020 & 0.020 & 0.018 & 0.023 & \textbf{0.019} & \textbf{0.019} \\
wine & 0.140 & 0.139 & 0.106 & 0.089 & 0.102 & 0.064 & 0.084 & 0.083 & 0.067 & \textbf{0.063} & \textbf{0.063} \\
collision-det & 0.142 & 0.137 & 0.088 & 0.059 & 0.084 & 0.032 & 0.041 & 0.041 & 0.036 & \textbf{0.030} & 0.031 \\
mnist-1-5 & 0.249 & 0.234 & 0.066 & 0.053 & 0.067 & 0.046 & 0.052 & 0.047 & 0.054 & \textbf{0.044} & \textbf{0.044} \\
mnist-2-6 & 0.268 & 0.253 & 0.066 & 0.054 & 0.063 & 0.047 & 0.048 & 0.046 & 0.053 & \textbf{0.045} & \textbf{0.045} \\
mnist & 0.395 & 0.381 & 0.118 & 0.109 & 0.112 & 0.065 & 0.081 & 0.077 & 0.076 & \textbf{0.061} & 0.062 \\
F-mnist2v5 & 0.273 & 0.247 & 0.230 & 0.188 & 0.234 & 0.165 & 0.173 & 0.165 & 0.189 & \textbf{0.156} & \textbf{0.156} \\
F-mnist3v4 & 0.290 & 0.269 & 0.225 & 0.196 & 0.220 & 0.146 & 0.176 & 0.167 & 0.179 & \textbf{0.135} & 0.136 \\
F-mnist7v9 & 0.283 & 0.280 & 0.234 & 0.205 & 0.226 & 0.156 & 0.168 & \textbf{0.144} & 0.193 & \textbf{0.144} & 0.145 \\
F-mnist & 0.427 & 0.401 & 0.283 & 0.235 & 0.259 & 0.121 & 0.171 & 0.153 & 0.153 & \textbf{0.110} & 0.109 \\
cifar10:0v5 & 0.419 & 0.380 & 0.305 & 0.231 & 0.287 & 0.137 & 0.173 & 0.166 & 0.156 & \textbf{0.130} & 0.129 \\
cifar10:0v6 & 0.403 & 0.372 & 0.329 & 0.278 & 0.328 & 0.204 & 0.214 & 0.212 & 0.222 & \textbf{0.198} & \textbf{0.198} \\
cifar10:4v8 & 0.408 & 0.392 & 0.326 & 0.275 & 0.310 & 0.208 & 0.211 & \textbf{0.203} & 0.215 & \textbf{0.205} & \textbf{0.205} \\ \hline
\textbf{AVERAGE} & 0.237 & 0.224 & 0.151 & 0.127 & 0.146 & 0.086 & 0.102 & 0.095 & 0.098 & \textbf{0.082} & \textbf{0.082} \\ \hline
\end{tabular}
}
\caption{
Averaged max regret for ensemble forests methods. The best results are bolded.}
\label{tab:DF-regret}
\end{table*}

\vspace{0.2em}\noindent\textbf{Ablation study.}
To assess the influence of specific ICoEvoRDF components, we conducted an ablation study. We tested ICoEvoRDF with an alternative voting method---\textit{equal voting (EV)}. Instead of calculating the mixed Nash equilibrium and using their probabilities as voting weights, EV assigns equal contribution to each island representative. This approach is common in classical random forest settings.

Another tested aspect of ICoEvoRDF was the input data selection. In the baseline version, a unique training set is assigned to each island by sampling with replacement. To evaluate the influence of this strategy, we tested an alternative approach: selecting the same input (SI) of all training examples as the training dataset $X$ for each island.
The results of these ablation studies are presented in Tables~\ref{tab:DF-accuracy} and~\ref{tab:DF-regret}. Even the simplified island-based model with EV and SI settings (ICoEvoRDF$\phantom{}^{EV}_{SI}$) outperformed the baseline CoEvoRDT forest method. Further improvements were observed with the incorporation of Nash voting (ICoEvoRDF$\phantom{}_{SI}$) and distinct input sets for each island (ICoEvoRDF$\phantom{}^{EV}$). The most significant improvement was achieved by combining NV with distinct input sets, i.e., ICoEvoRDF.

Additional experiments (presented later in this section) indicate that these improvements are correlated with increased diversity in the generated DT ensembles. Utilizing different inputs for each island enhances the diversity of the final ensemble, while Nash voting effectively assigns weights to these diverse DTs. This synergy between differentiated inputs and Nash voting leads to a significant improvement in performance.

\vspace{0.2em}\noindent\textbf{Computation time.}
Due to space limits, a detailed computation time comparison is provided in the supplementary material. Here, we only report the sum of the average computation times across all benchmarks and discuss the general conclusions.

Methods for creating single DTs (FPRDT and CoEvoRDT) exhibit significantly shorter average computation times (FPRDT: 246s, CoEvoRDT: 727s) compared to ensemble methods (PRAdaBoost: 6972s, CoEvoRDT boosting: 7606s, ICoEvoRDF: 6845s, ICoEvoRDF+FPRDT: 7091s). Computation times for all ensemble methods are of the same order of magnitude, ensuring a fair comparison across methods under a similar time constraint. The scalability and performance of ICoEvoRDF under increased computational budgets are presented in the supplementary material.

The results presented above do not utilize parallelization, which could substantially reduce the computation time for non-boosting ensemble methods (CoEvoRDT forest, ICoEvoRDF, and ICoEvoRDT + FPRDT). Parallelization is straightforward for the CoEvoRDT forest, as each run of the CoEvoRDT algorithm is independent and can be executed concurrently. However, migration in islanding methods complicates parallelization, although it can be facilitated by incorporating shared memory. Each island can store its $k_{\mathrm{top}}$ individuals from the last generation in this shared memory, and instead of a synchronous migration phase where each island sends its $k_{\mathrm{top}}$ individuals to neighbors, islands can retrieve the individuals needing migration from the shared memory. We have verified that this approach does not impact the results and reduces computation time by a factor of 8-10. A detailed discussion of this optimization and its results can be found in the supplementary material. For clarity of the ICoEvoRDT description, we have opted to omit the details of this optimization from the main body of the paper.
Such parallelization cannot be implemented for boosting methods (PRAdaBoost and CoEvoRDT boosting), which is an additional advantage of the proposed island-based approach.

\vspace{0.2em}\noindent\textbf{Diversity analysis.}
We hypothesize that the performance differences
between different variants of the proposed algorithm 
are related to the diversity of DTs generated by the algorithm. To quantify the diversity within a set of DTs, we measure the fraction of perturbed input instances for which the predictions of each pair of trees in the set differ.

Formally, let $\mathcal{T} = \{T_1, T_2, ..., T_k\}$ represent the ensemble of decision trees, and let $X' = \{x'_1, x'_2, ..., x'_m\}$ denote the set of perturbed input instances. The diversity between two trees $T_i$ and $T_j$ is defined as:
$\mathrm{div}(T_i, T_j) = \frac{1}{m} \sum_{k=1}^{m} I[T_i(x'_k) \neq T_j(x'_k)]$.

The \textit{average diversity} of the ensemble is then given by:
\begin{align}\mathrm{avg\_div}(\mathcal{T}) = \frac{2}{n(n-1)} \sum_{i=1}^{n-1} \sum_{j=i+1}^{n} \mathrm{div}(T_i, T_j)\end{align}
and the \textit{maximum diversity} is defined as:
\begin{align}\mathrm{max\_div}(\mathcal{T}) = \max_{1 \leq i < j \leq n} \mathrm{div}(T_i, T_j)\end{align}

We conducted a diversity analysis for two variants: \textit{external} and \textit{internal} diversity. External diversity was calculated using the decision trees that comprise the final ensemble, while internal diversity was assessed on the DTs within each island population.
The results of the diversity analysis are presented in Table~\ref{tab:diversity}. Given that the voting method is independent of diversity and does not influence its value, Nash voting and equal voting were 
considered jointly.

\begin{table}[ht]
\setlength{\tabcolsep}{2pt}
\renewcommand{\arraystretch}{1.05}
\resizebox{1.0\columnwidth}{!}{
\begin{tabular}{|c|c|c|c|c|}
\hline
diversity & $N$ CoEvoRDT & ICoEvoRDF$\phantom{}_{SI}$ & ICoEvoRDF & \begin{tabular}[c]{@{}c@{}} ICoEvoRDF\\ + FPRDT\end{tabular} \\ \hline
external avg & 0.037 & 0.038 & 0.045 & 0.054 \\ \hline
external max & 0.041 & 0.042 & 0.050 & 0.061 \\ \hline
internal avg & 0.107 & 0.110 & 0.111 & 0.109 \\ \hline
internal max & 0.200 & 0.211 & 0.209 & 0.210 \\ \hline
\end{tabular}
}
\caption{Diversity averaged over all benchmark datasets.}
\label{tab:diversity}
\end{table}

Our first observation is that internal is significantly higher than external diversity. This difference arises from the dynamic nature of the 
DT population evaluation, as each population potentially contains weaker DTs 
resulting from the exploratory nature of mutation operations, 
leading to higher internal diversity. Secondly, internal diversity appears to be consistent across different methods, while external diversity varies. The primary proposed method, ICoEvoRDF, achieves the highest average and maximum external diversity across all datasets when compared to the method without migration ($N$ ICoEvoRDF) and the method using the same training set for each island (ICoEvoRDF$\phantom{}_{SI}$). This suggests that the introduction of migration and varying inputs leads to a more diverse ensemble of 
DTs. Furthermore, incorporating FPRDT as an initialization for one island (ICoEvoRDF+FPRDT) further enhances diversity. These findings suggest a positive correlation between the diversity of the generated ensemble and the method's performance.
A more diverse ensemble is likely to be more robust against adversarial attacks and to generalize better to unseen data.

\vspace{0.2em}\noindent\textbf{Single decision trees.}
While ICoEvoRDF is designed for the construction of random decision forests, it can also be utilized to produce a single DT by selecting the most robust tree across all islands. This method will be referred to as ICoEvoRD\textbf{T}. It resembles the $N$ CoEvoRDT method proposed in~\cite{zychowski2024CoEvoRDT}, which involves multiple independent runs of CoEvoRDT followed by the selection of the best DT. However, the island model described in this paper introduces the crucial element of migration between islands, each running an independent version of CoEvoRDT. Our experiments indicate that this migration mechanism significantly improves results, outperforming both the baseline coevolutionary method CoEvoRDT and its simple rerun counterpart ($N$ CoEvoRDT), where the number of repetitions ($N$) is equivalent to the number of islands in ICoEvoRDT. This underscores the importance of the individual migration concept detailed in the ICoEvoRDF algorithm description section.

For all benchmarks, the incorporation of islanding with individual migration yielded superior outcomes (averaged over all benchmarks) -- adversarial accuracy (AA): 0.779, max regret (MR): 0.113, compared to the baseline method CoEvoRDT (AA: 0.767, MR: 0.131) and all other competitors: GROOT (AA: 0.718, MR: 0.158), FPRDT (AA: 0.765, MR: 0.155), $N$ CoEvoRDT (AA: 0.776, MR: 0.122). Detailed results are presented in the supplementary material.

\section{Conclusions}

In this paper, we introduced ICoEvoRDF, a novel island-based coevolutionary algorithm for constructing robust decision forests. Our approach leverages the strengths of multiple independently evolving populations of decision trees and perturbations, with periodic migration of top-performing individuals between islands. This strategy fosters diversity and promotes the exploration of a wider range of potential solutions, leading to more robust decision tree ensembles.

Our experimental results on 20 datasets demonstrate the effectiveness of ICoEvoRDF in optimizing both adversarial accuracy and minimax regret metrics. The algorithm consistently outperforms state-of-the-art methods, showcasing its ability to generate highly robust decision trees and forests. The flexibility of ICoEvoRDF allows for the integration of decision trees from various existing methods, offering a unified framework for combining diverse solutions.

A notable aspect of our work is the trade-off between model interpretability and robustness. ICoEvoRDF can generate more robust ensemble models, although their complexity and the associated weighting may reduce interpretability. Conversely, it can produce simpler single DTs that are less robust but easier to interpret. The balance between robustness and interpretability can be adjusted by the number of islands, allowing to control which aspect is prioritized.

To the best of our knowledge, this is the first study where mixed Nash equilibrium has been combined with an island-based evolutionary algorithm. Our 
results demonstrate that this synergy between coevolutionary methods and game theory is highly effective. We believe that it 
holds significant application potential also
in other domains, worth
further investigation.
Future work can also focus on extending the application of island-based coevolutionary algorithms to other related tasks in machine learning. One promising direction is investigating the use of ICoEvoRDF to ensure fairness in machine learning models. By incorporating fairness metrics into the objective function and evolving decision trees that minimize bias, we can potentially develop fairer and more equitable decision-making systems. Another direction of work can be using a similar setting for analysis of models properties (e.g., explainability) or constructing other machine learning models for which the target metric is not differentiable or it is challenging to compute an exact value (like robustness metrics presented in this paper).

\clearpage


\section*{Supplementary material}

# Cultivating Archipelago of Forests: Evolving Robust Decision Trees through Island Coevolution
# – Supplementary material –

## ICoEvoRDF pseudocode

---
Algorithm 1: ICoEvoRDF pseudocode.
---
1: **Input:**
2:   $X$ - training dataset
3:   $|I|$ - number of islands
4:   $n_g$ - number of generations per island evolution phase
5:   $k_{\text{top}}$ - number of top individuals to migrate
6:   $l_g$ - total number of generations limit
7:   $l_c$ - number of generations without improvement limit
8:   $\xi$ - robustness metric to optimize (e.g., adversarial accuracy or minimax regret)
9: **Initialize Islands:**
10: **for** each island $I \in \mathcal{I}$ **do**
11:   Initialize $I^T$ with random decision trees
12:   Initialize $I^P$ by sampling perturbations from $N_\epsilon(X)$
13:   Sample training subset for $I$ from $X$ with replacement
14: **end for**
15: **Evolve Islands:**
16: $g \leftarrow 0$
17: **while** $g < l_g$ **and** improvement within last $l_c$ generations **do**
18:   **for** each island $I \in \mathcal{I}$ **do**
19:     Evolve $I^T$ and $I^P$ for $n_g$ generations using CoEvoRDT algorithm
20:     Evaluate fitness of individuals in $I^T$ against $I^P$ and vice versa
21:   **end for**
22:   Migrate $k_{\text{top}}$ best DTs and perturbations from each neighbor in $\eta(I)$ to $I^T$ and $I^P$, resp
23:   $g \leftarrow g + n_g$
24: **end while**
25: **Construct Forest:**
26: Select best DTs from each island: $T_i^{\text{best}} = V_\xi^1(I_i^T)$ for $i \in I$
27: Compute mixed Nash equilibrium to get weights: $w_i$ for $i \in I$
28: **Return:** Weighted ensemble of $T_i^{\text{best}}$ with weights $w_i$
---

## Tested benchmarks

Table 1 presents 20 benchmark datasets used in method evaluation. All selected datasets are publicly available at https://www.openml.org and can be downloaded with *fetch_openml* function from the *sklearn* Python library.

| dataset | $\varepsilon$ | Instances | Features | Classes |
|---|---|---|---|---|
| ionos | 0.2 | 351 | 34 | 2 |
| breast | 0.3 | 683 | 9 | 2 |
| diabetes | 0.05 | 768 | 8 | 2 |
| bank | 0.1 | 1372 | 4 | 2 |
| Japan:3v4 | 0.1 | 3087 | 14 | 2 |
| spam | 0.05 | 4601 | 57 | 2 |
| GesDvP | 0.01 | 4838 | 32 | 2 |
| har1v2 | 0.1 | 3266 | 561 | 2 |
| wine | 0.1 | 6497 | 11 | 2 |
| collision-det | 0.1 | 33000 | 6 | 2 |
| mnist:1v5 | 0.3 | 13866 | 784 | 2 |
| mnist:2v6 | 0.3 | 13866 | 784 | 2 |
| mnist | 0.3 | 70000 | 784 | 10 |
| f-mnist:2v5 | 0.2 | 14000 | 784 | 2 |
| f-mnist:3v4 | 0.2 | 14000 | 784 | 2 |
| f-mnist:7v9 | 0.2 | 14000 | 784 | 2 |
| f-mnist | 0.2 | 70000 | 784 | 10 |
| cifar10:0v5 | 0.1 | 12000 | 3072 | 2 |
| cifar10:0v6 | 0.1 | 12000 | 3072 | 2 |
| cifar10:4v8 | 0.1 | 12000 | 3072 | 2 |

Table 1: Properties of the tested benchmark datasets.

## Results for single decision trees

Tables 2 and 3 provide a comparative analysis of various methods for generating single DTs. Across all benchmarks, the incorporation of islanding with individual migration (ICoEvoRDT and ICoEvoRDT + FPRDT) yielded superior outcomes compared to the baseline coevolutionary method CoEvoRDT and all other competitors. This suggests that the island-based mechanism further enhances robustness not only for DT ensembles but also for single DTs.

## Computation times

Table 4 presents a comparison of computation times for the most notable methods. For ICoEvoRDF, the difference in computation time between voting methods (equal voting and Nash voting) and island training subsets (same inputs and different inputs) is negligible, thus only a single column is dedicated to all these variants.

The methods designed for creating single decision trees (FPRDT and CoEvoRDT) exhibit significantly shorter computation times compared to the ensemble methods. This is expected as ensemble methods involve training multiple trees and potentially additional steps like boosting or island evolution. Among the ensemble methods, PRAdaBoost, CoEvoRDT boosting, ICoEvoRDF, and ICoEvoRDF + FPRDT have computation times within the same order of magnitude.

| dataset | CART | RIGDT-h | GROOT | FPRDT | CoEvoRDT | CoEvoRDT + FPRDT | N CoEvoRDT | ICoEvoRDT | ICoEvoRDT + FPRDT |
|---|---|---|---|---|---|---|---|---|---|
| ionos | 0.310 | 0.701 | 0.783 | 0.795 | 0.791 | 0.795 | 0.796 | 0.796 | **0.797** |
| breast | 0.250 | 0.838 | 0.874 | 0.876 | 0.885 | 0.889 | 0.889 | 0.889 | **0.890** |
| diabetes | 0.542 | 0.569 | 0.623 | 0.648 | 0.617 | 0.648 | 0.637 | 0.627 | **0.650** |
| bank | 0.633 | 0.468 | 0.541 | 0.658 | 0.657 | 0.663 | 0.667 | **0.672** | **0.672** |
| Japan3v4 | 0.576 | 0.564 | 0.584 | 0.667 | 0.665 | 0.668 | 0.678 | **0.688** | **0.688** |
| spam | 0.302 | 0.467 | 0.723 | 0.746 | 0.751 | 0.753 | 0.757 | **0.760** | **0.760** |
| GesDvP | 0.478 | 0.548 | 0.716 | 0.735 | 0.740 | 0.741 | 0.744 | 0.746 | **0.747** |
| har1v2 | 0.232 | 0.707 | 0.806 | 0.804 | 0.818 | 0.820 | 0.838 | **0.856** | **0.856** |
| wine | 0.620 | 0.474 | 0.637 | 0.674 | 0.688 | 0.692 | 0.695 | **0.698** | **0.698** |
| collision-det | 0.743 | 0.764 | 0.784 | 0.792 | 0.798 | 0.803 | 0.807 | **0.811** | **0.811** |
| mnist-1-5 | 0.921 | 0.957 | 0.954 | 0.966 | 0.964 | 0.969 | 0.971 | **0.973** | **0.973** |
| mnist-2-6 | 0.862 | 0.919 | 0.917 | 0.922 | 0.917 | 0.922 | 0.924 | **0.925** | **0.925** |
| mnist | 0.673 | 0.704 | 0.743 | 0.742 | 0.745 | 0.754 | 0.756 | 0.758 | **0.759** |
| F-mnist2v5 | 0.675 | 0.945 | 0.971 | 0.978 | 0.982 | 0.982 | 0.986 | **0.991** | **0.991** |
| F-mnist3v4 | 0.632 | 0.793 | 0.819 | 0.865 | 0.869 | 0.870 | 0.875 | 0.880 | **0.881** |
| F-mnist7v9 | 0.642 | 0.810 | 0.829 | 0.876 | 0.868 | **0.880** | **0.880** | **0.880** | **0.880** |
| F-mnist | 0.464 | 0.525 | 0.536 | 0.531 | 0.544 | 0.546 | 0.550 | **0.553** | **0.553** |
| cifar10:0v5 | 0.296 | 0.347 | 0.485 | 0.678 | 0.685 | 0.693 | 0.697 | 0.700 | **0.701** |
| cifar10:0v6 | 0.587 | 0.477 | 0.556 | 0.688 | 0.692 | 0.697 | 0.700 | **0.703** | **0.703** |
| cifar10:4v8 | 0.256 | 0.488 | 0.473 | 0.661 | 0.663 | 0.664 | 0.669 | **0.673** | **0.673** |
| **AVERAGE** | 0.535 | 0.653 | 0.718 | 0.765 | 0.767 | 0.772 | 0.776 | 0.779 | **0.781** |

Table 2: Averaged adversarial accuracies for single decision trees. The best results are bolded.

ICoEvoRDF parallelization (see Section ) performed on 10 parallel processes significantly decreased computation time by nearly a factor of 9.

## ICoEvoRDF parallelization

In the main paper, we presented a simplified, non-parallelized version of ICoEvoRDF. However, the algorithm can be further optimized by enabling parallel computations across all islands. Algorithm 2 provides pseudocode that incorporates parallelization to enhance the algorithm's efficiency. The key difference to the sequential version is the approach to the migration process.

The parallelized version introduces a shared memory ($S$) to facilitate migration in a parallel setting. After each island completes its evolution phase, it stores its top $k_{\text{top}}$ individuals in this shared memory. Subsequently, each island retrieves the required individuals from its neighbors directly from the shared memory, eliminating the need for explicit communication or synchronization between islands during migration. Thanks to this evolution of each island can happen concurrently, potentially leveraging multiple processing cores or machines. This is indicated the phrase "in parallel" added to for loop that evolves the islands. The generation counter $g$ is updated by $\frac{n_g}{|\mathcal{I}|}$ in line 25. This adjustment ensures that the total number of generations across all islands remains consistent with the sequential version, as each island now progresses $n_g$ generations in parallel.

## ICoEvoRDF parameterization

ICoEvoRDF parameterization process was performed on *cod-rna* dataset with 9 features, 2 classes, and 48565 instances. This dataset was not used later in experimental evaluation described in the main paper. The algorithm was ex-

---

**Algorithm 2: Parallel ICoEvoRDF pseudocode.**

1: **Input:**
2:     $X$ - training dataset
3:     $|I|$ - number of islands
4:     $n_g$ - number of generations per island evolution phase
5:     $k_{\text{top}}$ - number of top individuals to migrate
6:     $l_g$ - total number of generations limit
7:     $l_c$ - number of generations without improvement limit
8:     $\xi$ - robustness metric to optimize (e.g., adversarial accuracy or minimax regret)
9:     $S$ - shared memory
10: **Initialize Islands:**
11: **for** each island $I \in \mathcal{I}$ **in parallel do**
12:     Initialize $I^T$ with random decision trees
13:     Initialize $I^P$ by sampling perturbations from $N_\epsilon(X)$
14:     Sample training subset for $I$ from $X$ with replacement
15: **end for**
16: **Evolve Islands:**
17: $g \leftarrow 0$
18: **while** $g < l_g$ **and** improvement within last $l_c$ generations **do**
19:     **for** each island $I \in \mathcal{I}$ **in parallel do**
20:         Evolve $I^T$ and $I^P$ for $n_g$ generations using CoEvoRDT
21:         Evaluate fitness of individuals in $I^T$ against $I^P$ and vice versa
22:         Store $k_{\text{top}}$ best DTs and perturbations from $I$ in $S$
23:         Retrieve $k_{\text{top}}$ best individuals from neighbors in $\eta(I)$ from $S$ and add them to $I^T$ and $I^P$
24:     **end for**
25:     $g \leftarrow g + \frac{n_g}{|\mathcal{I}|}$
26: **end while**
27: **Construct Forest:**
28: Select best DTs from each island: $T_i^{\text{best}} = V_\xi^1(I_i^T)$ for $i \in I$
29: Compute mixed Nash equilibrium to get weights: $w_i$ for $i \in I$
30: **Return:** Weighted ensemble of $T_i^{\text{best}}$ with weights $w_i$

| dataset | CART | RIGDT-h | GROOT | FPRDT | CoEvoRDT | CoEvoRDT + FPRDT | N CoEvoRDT | ICoEvoRDT | ICoEvoRDT + FPRDT |
|---|---|---|---|---|---|---|---|---|---|
| ionos | 0.094 | 0.071 | 0.061 | 0.061 | 0.052 | 0.052 | 0.050 | **0.049** | **0.049** |
| breast | 0.103 | 0.069 | 0.059 | 0.057 | 0.049 | 0.049 | 0.048 | **0.047** | **0.047** |
| diabetes | 0.202 | 0.132 | 0.124 | 0.117 | 0.096 | 0.094 | 0.087 | **0.079** | 0.081 |
| bank | 0.186 | 0.108 | 0.090 | 0.089 | 0.076 | 0.076 | 0.069 | **0.062** | **0.062** |
| Japan3v4 | 0.107 | 0.083 | 0.067 | 0.066 | 0.062 | 0.061 | 0.052 | **0.043** | 0.044 |
| spam | 0.097 | 0.083 | 0.074 | 0.074 | 0.070 | 0.069 | 0.066 | **0.062** | 0.063 |
| GesDvP | 0.152 | 0.133 | 0.129 | 0.131 | 0.114 | 0.114 | 0.113 | **0.111** | **0.111** |
| har1v2 | 0.105 | 0.084 | 0.068 | 0.068 | 0.064 | 0.064 | 0.053 | **0.042** | **0.042** |
| wine | 0.140 | 0.127 | 0.111 | 0.109 | 0.090 | 0.090 | 0.087 | **0.084** | **0.084** |
| collision-det | 0.142 | 0.093 | 0.088 | 0.091 | 0.061 | 0.059 | 0.055 | **0.050** | 0.051 |
| mnist-1-5 | 0.249 | 0.076 | 0.071 | 0.067 | 0.055 | 0.055 | 0.053 | **0.052** | 0.052 |
| mnist-2-6 | 0.268 | 0.087 | 0.072 | 0.069 | 0.055 | 0.054 | 0.053 | **0.051** | 0.051 |
| mnist | 0.395 | 0.139 | 0.125 | 0.124 | 0.113 | 0.112 | 0.103 | **0.093** | 0.094 |
| F-mnist2v5 | 0.273 | 0.249 | 0.223 | 0.238 | 0.196 | 0.196 | 0.190 | **0.183** | **0.183** |
| F-mnist3v4 | 0.290 | 0.254 | 0.246 | 0.232 | 0.202 | 0.199 | 0.191 | **0.181** | 0.183 |
| F-mnist7v9 | 0.283 | 0.251 | 0.237 | 0.240 | 0.208 | 0.207 | 0.196 | **0.184** | 0.185 |
| F-mnist | 0.427 | 0.337 | 0.292 | 0.286 | 0.238 | 0.237 | 0.219 | **0.201** | **0.201** |
| cifar10:0v5 | 0.419 | 0.379 | 0.347 | 0.314 | 0.241 | 0.236 | 0.216 | **0.193** | 0.196 |
| cifar10:0v6 | 0.403 | 0.368 | 0.342 | 0.341 | 0.289 | 0.289 | 0.269 | **0.248** | 0.249 |
| cifar10:4v8 | 0.408 | 0.360 | 0.339 | 0.331 | 0.283 | 0.281 | 0.262 | **0.243** | 0.244 |
| **AVERAGE** | 0.237 | 0.174 | 0.158 | 0.155 | 0.131 | 0.130 | 0.122 | **0.113** | 0.114 |

Table 3: Averaged max regret for single decision trees. The best results are bolded.

ecuted 10000 times with parameter values set randomly, i.e. for each run each parameter was drawn uniformly from some predefined set of values:

- decision trees population size
  $N_T : \{10, 20, 50, 100, 200, 500, \mathbf{1000}\}$
- perturbations population size
  $N_P : \{100, 200, 500, 1000, 2000, 5000, \mathbf{10000}\}$
- number of consecutive generations for each population
  $l_c : \{1, 2, 5, 10, \mathbf{20}, 50, 100\}$
- the number of the best individuals from the decision trees population involved in the perturbations evaluation
  $N_{\text{top}} : \{1, 2, 5, 10, \mathbf{20}, 50, 100, 200\}$
- crossover probability
  $p_c : \{0.0, 0.1, 0.2, 0.3, 0.4, 0.5, 0.6, 0.7, \mathbf{0.8}, 0.9, 1.0\}$
- mutation probability
  $p_m : \{0.0, 0.1, 0.2, 0.3, 0.4, \mathbf{0.5}, 0.6, 0.7, 0.8, 0.9, 1.0\}$
- selection pressure
  $p_s : \{0.5, 0.6, 0.7, 0.8, \mathbf{0.9}, 1.0\}$
- HoF size
  $N_{\text{HoF}} : \{0, 10, 20, 50, 100, 200, \mathbf{500}\}$
- generations without improvement limit
  $l_c : \{5, 10, 20, \mathbf{50}, 100, 200\}$
- generations limit
  $l_g : \{100, 200, 500, 1000, \mathbf{2000}, \mathbf{5000}\}$

Best values (with the lowest average minimax regret across all runs) are bolded.

## Topologies

We evaluated four distinct migration topologies as illustrated in Figure 1: ring, cycle, star, and clique. These topologies differ in the number of neighbors each island has: in the ring topology, each island is connected to exactly 2 neighbors; in the directed cycle topology, each island has a single neighbor; in the clique topology, each island is connected to all other islands, resulting in $|\mathcal{I}| - 1$ neighbors; in the star topology, one central island is connected to all other islands, while these peripheral islands are connected only to the central one.

Table 5 presents the results, including averaged adversarial accuracy and maximum regret, for all tested topologies. The ring topology yielded the best performance, followed by the cycle and star topologies. The clique topology produced the poorest results, likely due to the excessive number of migrations, which effectively homogenized the population across islands. This high frequency of migration reduced the opportunity for niche development on individual islands, causing them to converge towards a single, less diverse population.

## Number of islands and population sizes

In all experiments presented in the main paper, we maintained the number of islands at 10 to ensure computational time comparable to state-of-the-art methods. However, it remains an open question how performance might improve with an increased number of islands. Additionally, increasing the number of decision trees within each island could further enhance results. Clearly more islands and a larger population of decision trees expand the search space, allowing for the evaluation of more candidate trees.

Tables 6 through 11 present results from experiments conducted on two datasets: diabetes and cifar-10:0v5. These results demonstrate that given more computation time (by increasing both the number of islands or the size of the decision tree population) can lead to performance improvements. Specifically, the data indicate that adding more islands yields

| dataset | FPRDT | CoEvoRDT | PRAdaBoost | CoEvoRDT boosting | ICoEvoRDF | ICoEvoRDF + FPRDT | ICoEvoRDF parallelized |
|---|---|---|---|---|---|---|---|
| ionos | 1 | 2 | 5 | 23 | 22 | 23 | 3 |
| breast | 1 | 2 | 6 | 20 | 20 | 21 | 3 |
| diabetes | 1 | 3 | 6 | 31 | 30 | 31 | 3 |
| bank | 2 | 6 | 8 | 67 | 49 | 51 | 6 |
| Japan3v4 | 3 | 9 | 13 | 106 | 95 | 98 | 12 |
| spam | 4 | 13 | 42 | 151 | 141 | 145 | 18 |
| GesDvP | 4 | 11 | 76 | 112 | 90 | 94 | 12 |
| har1v2 | 4 | 12 | 212 | 143 | 103 | 107 | 12 |
| wine | 6 | 2 | 23 | 20 | 20 | 26 | 2 |
| collision-det | 16 | 17 | 217 | 199 | 230 | 246 | 25 |
| mnist-1-5 | 8 | 22 | 354 | 229 | 207 | 215 | 26 |
| mnist-2-6 | 7 | 24 | 310 | 249 | 251 | 258 | 32 |
| mnist | 21 | 68 | 447 | 703 | 715 | 736 | 81 |
| f-mnist2v5 | 8 | 23 | 521 | 255 | 212 | 220 | 27 |
| f-mnist3v4 | 9 | 25 | 842 | 273 | 262 | 271 | 34 |
| f-mnist7v9 | 9 | 26 | 431 | 282 | 212 | 221 | 23 |
| f-mnist | 19 | 79 | 754 | 817 | 668 | 687 | 78 |
| cifar10:0v5 | 40 | 146 | 978 | 1409 | 1297 | 1337 | 147 |
| cifar10:0v6 | 42 | 126 | 896 | 1345 | 1233 | 1275 | 133 |
| cifar10:4v8 | 41 | 111 | 831 | 1172 | 987 | 1028 | 114 |
| **SUM** | **246** | **727** | **6972** | **7606** | **6845** | **7091** | **792** |

Table 4: Computation times (in seconds) for various methods on the benchmark datasets.

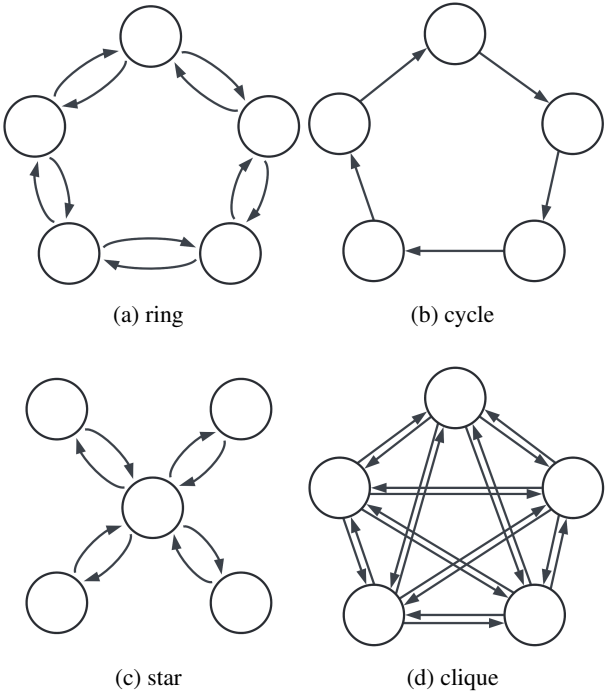

Figure 1: Tested islands migration topologies.

greater improvements compared to increasing the size of the decision tree population. Notably, no substantial gains are observed beyond 30 islands, suggesting diminishing returns with further increases in the number of islands.

## Metrics calculation

For calculating adversarial accuracy we used method based on Mixed Integer Linear Programming which calculate exact value of adversarial accuracy for decision trees.

However, calculating the exact value minimax regret is not straightforward. It requires finding a perturbation that maximizes regret from the infinite set of possible perturbations. Since this task is not trivial, we decided to estimate the real values of this metric by drawing a uniformly random subset $P$ of possible perturbations and then calculating the performance of all models on this subset.

In order to assess how large this subset should be to fairly estimate the performance of models, we chose 5 datasets with different $\varepsilon$ values and ran each tested method on each dataset 5 times. This resulted in 25 decision trees. We then checked the following values for the size of $P$: $10^2, 10^3, 10^4, 10^5, 10^6, 10^7$. For each value of $P$, we drew a given number of random perturbations and then evaluated all 25 decision trees using minimax regret and adversarial accuracy. The results for all models were then averaged. This procedure was repeated 20 times (each time a new subset of perturbations was drawn, but the 25 models remained the same) for each value of P. The mean value and standard error for the tested values of $P$ are presented in Table 12. It shows that the standard error value decreases with the size of $P$. This is expected, as a larger subset of perturbations allows for a more thorough search of the space of possible perturbations and indicates that the results are becoming more reliable.

|  | adversarial accuracy | | | | max regret | | | |
|---|---|---|---|---|---|---|---|---|
| dataset | ring | cycle | star | clique | ring | cycle | star | clique |
| ionos | 0.799 | 0.798 | 0.797 | 0.790 | 0.044 | 0.046 | 0.047 | 0.049 |
| breast | 0.900 | 0.899 | 0.894 | 0.882 | 0.034 | 0.034 | 0.034 | 0.035 |
| diabetes | 0.647 | 0.643 | 0.647 | 0.638 | 0.026 | 0.027 | 0.027 | 0.028 |
| bank | 0.673 | 0.670 | 0.667 | 0.664 | 0.046 | 0.047 | 0.048 | 0.051 |
| Japan3v4 | 0.688 | 0.684 | 0.687 | 0.679 | 0.025 | 0.026 | 0.027 | 0.027 |
| spam | 0.766 | 0.765 | 0.760 | 0.754 | 0.044 | 0.044 | 0.044 | 0.045 |
| GesDvP | 0.752 | 0.749 | 0.747 | 0.741 | 0.073 | 0.074 | 0.076 | 0.077 |
| har1v2 | 0.854 | 0.846 | 0.851 | 0.839 | 0.019 | 0.019 | 0.020 | 0.020 |
| wine | 0.708 | 0.703 | 0.704 | 0.694 | 0.063 | 0.065 | 0.066 | 0.067 |
| collision-detection | 0.822 | 0.818 | 0.822 | 0.812 | 0.030 | 0.031 | 0.032 | 0.032 |
| mnist-1-5 | 0.975 | 0.973 | 0.968 | 0.959 | 0.044 | 0.045 | 0.045 | 0.045 |
| mnist-2-6 | 0.925 | 0.918 | 0.922 | 0.915 | 0.045 | 0.045 | 0.046 | 0.047 |
| mnist | 0.764 | 0.760 | 0.763 | 0.752 | 0.061 | 0.063 | 0.065 | 0.067 |
| F-mnist2v5 | 0.995 | 0.991 | 0.990 | 0.982 | 0.156 | 0.158 | 0.164 | 0.174 |
| F-mnist3v4 | 0.884 | 0.877 | 0.883 | 0.870 | 0.135 | 0.135 | 0.138 | 0.139 |
| F-mnist7v9 | 0.881 | 0.880 | 0.877 | 0.866 | 0.144 | 0.146 | 0.147 | 0.155 |
| F-mnist | 0.560 | 0.557 | 0.560 | 0.549 | 0.110 | 0.113 | 0.115 | 0.119 |
| cifar10:0v5 | 0.702 | 0.699 | 0.702 | 0.689 | 0.130 | 0.135 | 0.138 | 0.145 |
| cifar10:0v6 | 0.704 | 0.703 | 0.701 | 0.694 | 0.198 | 0.202 | 0.204 | 0.210 |
| cifar10:4v8 | 0.675 | 0.670 | 0.674 | 0.664 | 0.205 | 0.211 | 0.215 | 0.225 |
| **AVERAGE** | **0.784** | **0.780** | **0.781** | **0.772** | **0.082** | **0.083** | **0.085** | **0.088** |

Table 5: Averaged adversarial accuracy and max regret for various migration topologies.

|  | minimax regret | |
|---|---|---|
| $log_{10}|P|$ | average | std error |
| 2 | 0.0954 | 0.0062 |
| 3 | 0.0963 | 0.0051 |
| 4 | 0.0970 | 0.0023 |
| 5 | 0.0972 | 0.0004 |
| 6 | 0.0973 | 0.0002 |
| 7 | 0.0973 | 0.0001 |

Table 12: Mean value and standard error of minimax regret for different values of the size of random perturbations sample used to their calculation.

The standard error for small values ($\log_{10}|P| \leq 4$) is high, which shows that the calculated metrics values are unreliable. However, for $P$ sizes of at least $10^5$, the difference between multiple perturbations drawn is small and the results are stabilized. This does not indicate how close to the exact (real) values we are, but it does show that $10^5$ is a large enough size of drawn perturbations sample to fairly assess and compare tested models.

### Diversity analysis

Table 13 presents detailed results for diversity analysis described in the main paper.

### Standard deviations

To ensure clarity and due to space constraints, we present the standard deviations of the results discussed in the main paper in separate tables. Tables 14 and 15 show results for decision forests while Tables 16 and 17 the corresponding provide results for individual decision trees.

The standard deviations for ICoEvoRDF don't significantly differ than those of other methods. This suggests that ICoEvoRDF produces consistent and stable results across different runs or evaluations. The standard deviations for single decision trees are notably higher than those for ensemble forests. This is expected, as ensembles tend to reduce variance and improve robustness compared to individual models. Results vary across datasets, indicating that some datasets are inherently more challenging or exhibit greater variability in performance, regardless of the method used.

|  |  | number of islands |  |  |  |  |  |  |  |  |  | |
|---|---|---|---|---|---|---|---|---|---|---|---|---|
|  |  | 5 | 10 | 15 | 20 | 25 | 30 | 35 | 40 | 45 | 50 | AVG |
| DT population size | 50 | 0.604 | 0.609 | 0.613 | 0.616 | 0.614 | 0.622 | 0.622 | 0.622 | 0.622 | 0.622 | 0.617 |
|  | 100 | 0.628 | 0.628 | 0.638 | 0.639 | 0.639 | 0.644 | 0.644 | 0.644 | 0.644 | 0.644 | 0.639 |
|  | 150 | 0.636 | 0.640 | 0.644 | 0.648 | 0.650 | 0.652 | 0.652 | 0.652 | 0.652 | 0.652 | 0.648 |
|  | 200 | 0.637 | 0.645 | 0.645 | 0.649 | 0.651 | 0.653 | 0.653 | 0.653 | 0.653 | 0.654 | 0.649 |
|  | 250 | 0.637 | 0.646 | 0.647 | 0.648 | 0.652 | 0.654 | 0.654 | 0.654 | 0.654 | 0.654 | 0.650 |
|  | 300 | 0.638 | 0.646 | 0.647 | 0.649 | 0.653 | 0.655 | 0.655 | 0.655 | 0.655 | 0.656 | 0.651 |
|  | 350 | 0.638 | 0.646 | 0.648 | 0.649 | 0.653 | 0.655 | 0.655 | 0.656 | 0.656 | 0.656 | 0.651 |
|  | 400 | 0.639 | 0.646 | 0.648 | 0.650 | 0.653 | 0.655 | 0.655 | 0.655 | 0.656 | 0.656 | 0.651 |
|  | 450 | 0.639 | 0.646 | 0.648 | 0.650 | 0.654 | 0.655 | 0.655 | 0.655 | 0.656 | 0.656 | 0.651 |
|  | 500 | 0.639 | 0.646 | 0.648 | 0.650 | 0.654 | 0.655 | 0.656 | 0.656 | 0.656 | 0.656 | 0.651 |
|  | AVG | 0.634 | 0.640 | 0.643 | 0.645 | 0.647 | 0.650 | 0.650 | 0.650 | 0.650 | 0.651 |  |

Table 6: Averaged adversarial accuracy for number of islands (columns) and DT populations size (rows) for **diabetes** dataset.

|  |  | number of islands |  |  |  |  |  |  |  |  |  | |
|---|---|---|---|---|---|---|---|---|---|---|---|---|
|  |  | 5 | 10 | 15 | 20 | 25 | 30 | 35 | 40 | 45 | 50 | AVG |
| DT population size | 50 | 0.028 | 0.028 | 0.028 | 0.028 | 0.028 | 0.028 | 0.028 | 0.028 | 0.028 | 0.028 | 0.028 |
|  | 100 | 0.027 | 0.027 | 0.027 | 0.027 | 0.027 | 0.027 | 0.027 | 0.027 | 0.027 | 0.027 | 0.027 |
|  | 150 | 0.027 | 0.027 | 0.027 | 0.027 | 0.026 | 0.026 | 0.026 | 0.026 | 0.026 | 0.026 | 0.027 |
|  | 200 | 0.027 | 0.027 | 0.027 | 0.027 | 0.026 | 0.026 | 0.026 | 0.026 | 0.026 | 0.026 | 0.027 |
|  | 250 | 0.027 | 0.027 | 0.027 | 0.027 | 0.026 | 0.026 | 0.026 | 0.026 | 0.026 | 0.026 | 0.026 |
|  | 300 | 0.027 | 0.027 | 0.027 | 0.026 | 0.026 | 0.026 | 0.026 | 0.026 | 0.026 | 0.026 | 0.026 |
|  | 350 | 0.027 | 0.027 | 0.027 | 0.026 | 0.026 | 0.026 | 0.026 | 0.026 | 0.026 | 0.026 | 0.026 |
|  | 400 | 0.027 | 0.027 | 0.027 | 0.026 | 0.026 | 0.026 | 0.026 | 0.026 | 0.026 | 0.026 | 0.026 |
|  | 450 | 0.027 | 0.027 | 0.027 | 0.026 | 0.026 | 0.026 | 0.026 | 0.026 | 0.026 | 0.026 | 0.026 |
|  | 500 | 0.027 | 0.027 | 0.027 | 0.026 | 0.026 | 0.026 | 0.026 | 0.026 | 0.026 | 0.026 | 0.026 |
|  | AVG | 0.027 | 0.027 | 0.027 | 0.027 | 0.027 | 0.026 | 0.026 | 0.026 | 0.026 | 0.026 |  |

Table 7: Averaged max regret for number of islands (columns) and DT populations size (rows) for **diabetes** dataset.

|  |  | number of islands |  |  |  |  |  |  |  |  |  | |
|---|---|---|---|---|---|---|---|---|---|---|---|---|
|  |  | 5 | 10 | 15 | 20 | 25 | 30 | 35 | 40 | 45 | 50 | AVG |
| DT population size | 50 | 5 | 10 | 15 | 19 | 24 | 28 | 34 | 39 | 44 | 48 | 27 |
|  | 100 | 10 | 18 | 29 | 38 | 46 | 54 | 64 | 75 | 82 | 98 | 51 |
|  | 150 | 14 | 27 | 44 | 59 | 72 | 83 | 104 | 115 | 126 | 138 | 78 |
|  | 200 | 19 | 37 | 55 | 77 | 99 | 120 | 131 | 151 | 175 | 194 | 106 |
|  | 250 | 24 | 48 | 73 | 95 | 122 | 143 | 174 | 192 | 207 | 240 | 132 |
|  | 300 | 29 | 60 | 83 | 109 | 136 | 176 | 197 | 227 | 255 | 289 | 156 |
|  | 350 | 32 | 68 | 100 | 135 | 167 | 203 | 224 | 265 | 297 | 347 | 184 |
|  | 400 | 38 | 79 | 109 | 146 | 194 | 223 | 271 | 310 | 345 | 393 | 211 |
|  | 450 | 42 | 86 | 128 | 176 | 204 | 263 | 298 | 331 | 378 | 425 | 233 |
|  | 500 | 47 | 93 | 148 | 195 | 227 | 277 | 333 | 376 | 409 | 490 | 259 |
|  | AVG | 26 | 53 | 78 | 105 | 129 | 157 | 183 | 208 | 232 | 266 |  |

Table 8: Averaged computation time (in seconds) for number of islands (columns) and DT populations size (rows) for **diabetes** dataset.

|  | | number of islands | | | | | | | | | | AVG |
|---|---|---|---|---|---|---|---|---|---|---|---|---|
|  | | 5 | 10 | 15 | 20 | 25 | 30 | 35 | 40 | 45 | 50 | |
| DT population size | 50  | 0.660 | 0.663 | 0.666 | 0.664 | 0.670 | 0.668 | 0.666 | 0.666 | 0.667 | 0.667 | 0.666 |
| | 100 | 0.681 | 0.684 | 0.689 | 0.685 | 0.692 | 0.690 | 0.693 | 0.693 | 0.693 | 0.693 | 0.689 |
| | 150 | 0.694 | 0.695 | 0.698 | 0.699 | 0.700 | 0.702 | 0.701 | 0.702 | 0.702 | 0.702 | 0.699 |
| | 200 | 0.695 | 0.697 | 0.699 | 0.701 | 0.702 | 0.703 | 0.703 | 0.704 | 0.704 | 0.705 | 0.701 |
| | 250 | 0.696 | 0.698 | 0.700 | 0.701 | 0.702 | 0.702 | 0.705 | 0.705 | 0.706 | 0.706 | 0.702 |
| | 300 | 0.697 | 0.697 | 0.700 | 0.701 | 0.701 | 0.703 | 0.705 | 0.706 | 0.706 | 0.706 | 0.702 |
| | 350 | 0.697 | 0.697 | 0.700 | 0.701 | 0.702 | 0.703 | 0.706 | 0.706 | 0.706 | 0.706 | 0.702 |
| | 400 | 0.698 | 0.698 | 0.700 | 0.702 | 0.703 | 0.703 | 0.706 | 0.706 | 0.707 | 0.707 | 0.703 |
| | 450 | 0.698 | 0.698 | 0.700 | 0.702 | 0.703 | 0.704 | 0.706 | 0.707 | 0.707 | 0.707 | 0.703 |
| | 500 | 0.698 | 0.698 | 0.700 | 0.702 | 0.703 | 0.704 | 0.706 | 0.707 | 0.707 | 0.707 | 0.703 |
| | AVG | 0.691 | 0.692 | 0.695 | 0.696 | 0.698 | 0.698 | 0.700 | 0.700 | 0.701 | 0.701 | |

Table 9: Averaged adversarial accuracy for number of islands (columns) and DT populations size (rows) for **cifar10:0v5** dataset.

|  | | number of islands | | | | | | | | | | AVG |
|---|---|---|---|---|---|---|---|---|---|---|---|---|
|  | | 5 | 10 | 15 | 20 | 25 | 30 | 35 | 40 | 45 | 50 | |
| DT population size | 50  | 0.139 | 0.138 | 0.138 | 0.138 | 0.137 | 0.137 | 0.138 | 0.138 | 0.138 | 0.137 | 0.138 |
| | 100 | 0.135 | 0.134 | 0.133 | 0.134 | 0.133 | 0.133 | 0.132 | 0.132 | 0.132 | 0.132 | 0.133 |
| | 150 | 0.132 | 0.132 | 0.132 | 0.131 | 0.131 | 0.131 | 0.131 | 0.131 | 0.131 | 0.131 | 0.131 |
| | 200 | 0.132 | 0.132 | 0.131 | 0.131 | 0.131 | 0.131 | 0.130 | 0.130 | 0.130 | 0.130 | 0.131 |
| | 250 | 0.132 | 0.131 | 0.131 | 0.131 | 0.131 | 0.131 | 0.130 | 0.130 | 0.130 | 0.130 | 0.131 |
| | 300 | 0.132 | 0.132 | 0.131 | 0.131 | 0.131 | 0.131 | 0.130 | 0.130 | 0.130 | 0.130 | 0.131 |
| | 350 | 0.132 | 0.132 | 0.131 | 0.131 | 0.131 | 0.130 | 0.130 | 0.130 | 0.130 | 0.130 | 0.131 |
| | 400 | 0.131 | 0.131 | 0.131 | 0.131 | 0.131 | 0.130 | 0.130 | 0.130 | 0.130 | 0.130 | 0.131 |
| | 450 | 0.131 | 0.131 | 0.131 | 0.131 | 0.131 | 0.130 | 0.130 | 0.130 | 0.130 | 0.130 | 0.131 |
| | 500 | 0.131 | 0.131 | 0.131 | 0.131 | 0.131 | 0.130 | 0.130 | 0.130 | 0.130 | 0.130 | 0.130 |
| | AVG | 0.133 | 0.133 | 0.132 | 0.132 | 0.132 | 0.131 | 0.131 | 0.131 | 0.131 | 0.131 | |

Table 10: Averaged max regret for number of islands (columns) and DT populations size (rows) for **cifar10:0v5** dataset.

|  | | number of islands | | | | | | | | | | AVG |
|---|---|---|---|---|---|---|---|---|---|---|---|---|
|  | | 5 | 10 | 15 | 20 | 25 | 30 | 35 | 40 | 45 | 50 | |
| DT population size | 50  | 208 | 427 | 648 | 856 | 990 | 1265 | 1501 | 1587 | 1911 | 2000 | 1139 |
| | 100 | 395 | 809 | 1280 | 1657 | 2034 | 2548 | 2983 | 3392 | 3884 | 4033 | 2301 |
| | 150 | 608 | 1191 | 1927 | 2573 | 3178 | 3669 | 4191 | 5037 | 5318 | 6395 | 3409 |
| | 200 | 793 | 1727 | 2550 | 3301 | 4317 | 4812 | 6056 | 6874 | 7349 | 8518 | 4630 |
| | 250 | 985 | 2014 | 3228 | 3978 | 5218 | 6377 | 6985 | 8573 | 8877 | 10670 | 5690 |
| | 300 | 1214 | 2452 | 3605 | 5036 | 6053 | 7668 | 8948 | 10133 | 10766 | 12829 | 6870 |
| | 350 | 1380 | 2952 | 4369 | 5789 | 7090 | 9106 | 10489 | 11991 | 13646 | 13937 | 8075 |
| | 400 | 1693 | 3208 | 5003 | 6457 | 8659 | 10029 | 11418 | 13624 | 15268 | 16790 | 9215 |
| | 450 | 1853 | 3604 | 5571 | 7137 | 9226 | 10615 | 12580 | 14727 | 17237 | 18333 | 10088 |
| | 500 | 2149 | 4003 | 6448 | 8293 | 10197 | 12500 | 13814 | 16045 | 17910 | 20545 | 11190 |
| | AVG | 1128 | 2239 | 3463 | 4508 | 5696 | 6859 | 7897 | 9198 | 10217 | 11405 | |

Table 11: Averaged computation times (in seconds) for number of islands (columns) and DT populations size (rows) for **cifar10:0v5** dataset.

|  | external diversity | | | | | | | | internal diversity | | | | | | | |
|---|---|---|---|---|---|---|---|---|---|---|---|---|---|---|---|---|
|  | N CoEvoRDT | | ICoEvoRDF$_{SI}$ | | ICoEvoRDF | | ICoEvoRDF + FPRDT | | N CoEvoRDT | | ICoEvoRDF$_{SI}$ | | ICoEvoRDF | | ICoEvoRDF + FPRDT | |
| dataset | avg | max | avg | max | avg | max | avg | max | avg | max | avg | max | avg | max | avg | max |
| ionos | 0.037 | 0.040 | 0.038 | 0.041 | 0.046 | 0.050 | 0.055 | 0.064 | 0.119 | 0.222 | 0.128 | 0.25 | 0.126 | 0.237 | 0.127 | 0.238 |
| breast | 0.028 | 0.033 | 0.028 | 0.030 | 0.035 | 0.039 | 0.042 | 0.049 | 0.114 | 0.224 | 0.117 | 0.226 | 0.115 | 0.227 | 0.113 | 0.218 |
| diabetes | 0.050 | 0.056 | 0.051 | 0.056 | 0.062 | 0.069 | 0.063 | 0.077 | 0.098 | 0.19 | 0.101 | 0.184 | 0.106 | 0.191 | 0.104 | 0.192 |
| bank | 0.026 | 0.032 | 0.027 | 0.028 | 0.033 | 0.036 | 0.042 | 0.043 | 0.106 | 0.208 | 0.11 | 0.216 | 0.107 | 0.195 | 0.112 | 0.202 |
| Japan3v4 | 0.049 | 0.058 | 0.051 | 0.057 | 0.061 | 0.066 | 0.079 | 0.082 | 0.093 | 0.17 | 0.099 | 0.186 | 0.099 | 0.188 | 0.092 | 0.178 |
| spam | 0.036 | 0.041 | 0.038 | 0.044 | 0.044 | 0.047 | 0.052 | 0.060 | 0.127 | 0.24 | 0.132 | 0.26 | 0.128 | 0.248 | 0.132 | 0.241 |
| GesDvP | 0.057 | 0.060 | 0.054 | 0.063 | 0.063 | 0.072 | 0.065 | 0.086 | 0.103 | 0.19 | 0.106 | 0.194 | 0.104 | 0.204 | 0.102 | 0.201 |
| har1v2 | 0.039 | 0.043 | 0.039 | 0.045 | 0.046 | 0.048 | 0.056 | 0.061 | 0.109 | 0.2 | 0.108 | 0.199 | 0.116 | 0.211 | 0.116 | 0.232 |
| wine | 0.045 | 0.053 | 0.047 | 0.055 | 0.055 | 0.063 | 0.069 | 0.078 | 0.113 | 0.216 | 0.113 | 0.226 | 0.12 | 0.227 | 0.113 | 0.224 |
| collision-det | 0.028 | 0.033 | 0.031 | 0.036 | 0.036 | 0.036 | 0.043 | 0.045 | 0.127 | 0.229 | 0.13 | 0.246 | 0.135 | 0.246 | 0.137 | 0.246 |
| mnist-1-5 | 0.028 | 0.031 | 0.030 | 0.033 | 0.036 | 0.041 | 0.038 | 0.050 | 0.100 | 0.18 | 0.105 | 0.189 | 0.102 | 0.186 | 0.107 | 0.195 |
| mnist-2-6 | 0.025 | 0.028 | 0.026 | 0.026 | 0.032 | 0.032 | 0.038 | 0.044 | 0.097 | 0.178 | 0.103 | 0.196 | 0.097 | 0.184 | 0.099 | 0.188 |
| mnist | 0.049 | 0.058 | 0.052 | 0.053 | 0.062 | 0.070 | 0.079 | 0.081 | 0.102 | 0.194 | 0.103 | 0.2 | 0.104 | 0.192 | 0.104 | 0.2 |
| F-mnist2v5 | 0.037 | 0.042 | 0.038 | 0.039 | 0.048 | 0.050 | 0.059 | 0.067 | 0.116 | 0.209 | 0.118 | 0.229 | 0.115 | 0.228 | 0.122 | 0.237 |
| F-mnist3v4 | 0.043 | 0.045 | 0.043 | 0.047 | 0.053 | 0.062 | 0.066 | 0.069 | 0.098 | 0.185 | 0.096 | 0.18 | 0.097 | 0.188 | 0.102 | 0.19 |
| F-mnist7v9 | 0.027 | 0.028 | 0.028 | 0.031 | 0.034 | 0.035 | 0.042 | 0.049 | 0.110 | 0.202 | 0.112 | 0.213 | 0.109 | 0.205 | 0.111 | 0.202 |
| F-mnist | 0.022 | 0.022 | 0.023 | 0.025 | 0.028 | 0.032 | 0.035 | 0.041 | 0.097 | 0.189 | 0.1 | 0.189 | 0.101 | 0.201 | 0.101 | 0.184 |
| cifar10:0v5 | 0.039 | 0.044 | 0.041 | 0.045 | 0.043 | 0.047 | 0.045 | 0.050 | 0.107 | 0.196 | 0.105 | 0.198 | 0.114 | 0.22 | 0.112 | 0.221 |
| cifar10:0v6 | 0.034 | 0.035 | 0.036 | 0.039 | 0.042 | 0.048 | 0.054 | 0.061 | 0.093 | 0.172 | 0.097 | 0.188 | 0.098 | 0.189 | 0.093 | 0.179 |
| cifar10:4v8 | 0.040 | 0.047 | 0.039 | 0.040 | 0.045 | 0.051 | 0.058 | 0.069 | 0.107 | 0.2 | 0.11 | 0.209 | 0.11 | 0.208 | 0.11 | 0.208 |
| **AVERAGE** | 0.037 | 0.041 | 0.038 | 0.042 | 0.045 | 0.050 | 0.054 | 0.061 | 0.107 | 0.200 | 0.110 | 0.211 | 0.111 | 0.209 | 0.109 | 0.210 |

Table 13: Diversity analysis in terms of average and maximum diversity.

| dataset | Random forests | GROOT forests | FPRDT forest | CoEvoRDT forest | PRAdaBoost | CoEvoRDT boosting | ICoEvoRDF$_{SI}^{EV}$ | ICoEvoRDF$_{SI}$ | ICoEvoRDF$^{EV}$ | ICoEvoRDF | ICoEvoRDF + FPRDT |
|---|---|---|---|---|---|---|---|---|---|---|---|
| ionos | 0.0028 | 0.0058 | 0.0048 | 0.0071 | 0.0080 | 0.0043 | 0.0070 | 0.0056 | 0.0070 | 0.0144 | 0.0130 |
| breast | 0.0106 | 0.0106 | 0.0099 | 0.0097 | 0.0114 | 0.0097 | 0.0093 | 0.0058 | 0.0122 | 0.0062 | 0.0057 |
| diabetes | 0.0059 | 0.0068 | 0.0067 | 0.0108 | 0.0125 | 0.0054 | 0.0050 | 0.0105 | 0.0135 | 0.0086 | 0.0081 |
| bank | 0.0061 | 0.0075 | 0.0065 | 0.0081 | 0.0098 | 0.0071 | 0.0109 | 0.0095 | 0.0088 | 0.0105 | 0.0098 |
| Japan3v4 | 0.0037 | 0.0073 | 0.0060 | 0.0036 | 0.0041 | 0.0062 | 0.0083 | 0.0050 | 0.0112 | 0.0071 | 0.0066 |
| spam | 0.0000 | 0.0079 | 0.0074 | 0.0028 | 0.0031 | 0.0062 | 0.0103 | 0.0105 | 0.0078 | 0.0101 | 0.0092 |
| GesDvP | 0.0098 | 0.0072 | 0.0062 | 0.0098 | 0.0111 | 0.0060 | 0.0111 | 0.0083 | 0.0086 | 0.0133 | 0.0129 |
| har1v2 | 0.0066 | 0.0069 | 0.0060 | 0.0108 | 0.0118 | 0.0066 | 0.0115 | 0.0076 | 0.0060 | 0.0084 | 0.0078 |
| wine | 0.0081 | 0.0090 | 0.0087 | 0.0076 | 0.0089 | 0.0091 | 0.0064 | 0.0069 | 0.0143 | 0.0081 | 0.0080 |
| collision-det | 0.0054 | 0.0111 | 0.0094 | 0.0045 | 0.0051 | 0.0079 | 0.0080 | 0.0125 | 0.0136 | 0.0125 | 0.0118 |
| mnist-1-5 | 0.0000 | 0.0081 | 0.0072 | 0.0068 | 0.0083 | 0.0073 | 0.0115 | 0.0059 | 0.0103 | 0.0100 | 0.0097 |
| mnist-2-6 | 0.0000 | 0.0023 | 0.0021 | 0.0089 | 0.0090 | 0.0019 | 0.0112 | 0.0084 | 0.0097 | 0.0090 | 0.0089 |
| mnist | 0.0000 | 0.0072 | 0.0068 | 0.0093 | 0.0108 | 0.0063 | 0.0068 | 0.0120 | 0.0065 | 0.0068 | 0.0066 |
| F-mnist2v5 | 0.0047 | 0.0080 | 0.0077 | 0.0073 | 0.0074 | 0.0083 | 0.0086 | 0.0125 | 0.0079 | 0.0132 | 0.0131 |
| F-mnist3v4 | 0.0023 | 0.0063 | 0.0058 | 0.0112 | 0.0112 | 0.0057 | 0.0128 | 0.0052 | 0.0144 | 0.0093 | 0.0088 |
| F-mnist7v9 | 0.0050 | 0.0078 | 0.0077 | 0.0033 | 0.0035 | 0.0062 | 0.0095 | 0.0083 | 0.0070 | 0.0138 | 0.0127 |
| F-mnist | 0.0050 | 0.0044 | 0.0043 | 0.0078 | 0.0080 | 0.0036 | 0.0066 | 0.0129 | 0.0089 | 0.0072 | 0.0066 |
| cifar10:0v5 | 0.0094 | 0.0074 | 0.0074 | 0.0094 | 0.0101 | 0.0064 | 0.0123 | 0.0067 | 0.0092 | 0.0128 | 0.0119 |
| cifar10:0v6 | 0.0088 | 0.0030 | 0.0028 | 0.0033 | 0.0039 | 0.0032 | 0.0068 | 0.0072 | 0.0063 | 0.0114 | 0.0109 |
| cifar10:4v8 | 0.0059 | 0.0087 | 0.0073 | 0.0109 | 0.0130 | 0.0080 | 0.0108 | 0.0114 | 0.0065 | 0.0087 | 0.0080 |
| **AVERAGE** | 0.0050 | 0.0072 | 0.0065 | 0.0077 | 0.0085 | 0.0063 | 0.0092 | 0.0086 | 0.0095 | 0.0101 | 0.0095 |

Table 14: Standard deviations of adversarial accuracies for ensemble forests methods.

| dataset | Random forests | GROOT forests | FPRDT forest | CoEvoRDT forest | PRAdaBoost | CoEvoRDT boosting | ICoEvoRDF$_{SI}^{EV}$ | ICoEvoRDF$_{SI}$ | ICoEvoRDF$^{EV}$ | ICoEvoRDF | ICoEvoRDF + FPRDT |
|---|---|---|---|---|---|---|---|---|---|---|---|
| breast | 0.0034 | 0.0034 | 0.0075 | 0.0050 | 0.0061 | 0.0078 | 0.0058 | 0.0019 | 0.0072 | 0.0046 | 0.0042 |
| diabetes | 0.0019 | 0.0030 | 0.0048 | 0.0077 | 0.0071 | 0.0030 | 0.0036 | 0.0036 | 0.0078 | 0.0050 | 0.0037 |
| bank | 0.0028 | 0.0030 | 0.0047 | 0.0055 | 0.0038 | 0.0039 | 0.0054 | 0.0045 | 0.0061 | 0.0073 | 0.0044 |
| Japan3v4 | 0.0013 | 0.0038 | 0.0041 | 0.0022 | 0.0016 | 0.0047 | 0.0030 | 0.0024 | 0.0050 | 0.0049 | 0.0037 |
| spam | 0.0000 | 0.0051 | 0.0050 | 0.0019 | 0.0023 | 0.0024 | 0.0064 | 0.0079 | 0.0045 | 0.0080 | 0.0035 |
| GesDvP | 0.0067 | 0.0044 | 0.0049 | 0.0059 | 0.0069 | 0.0047 | 0.0070 | 0.0063 | 0.0046 | 0.0104 | 0.0088 |
| har1v2 | 0.0032 | 0.0024 | 0.0040 | 0.0044 | 0.0060 | 0.0047 | 0.0035 | 0.0052 | 0.0025 | 0.0044 | 0.0038 |
| wine | 0.0033 | 0.0047 | 0.0053 | 0.0049 | 0.0030 | 0.0072 | 0.0047 | 0.0024 | 0.0096 | 0.0044 | 0.0054 |
| collision-det | 0.0042 | 0.0044 | 0.0075 | 0.0025 | 0.0016 | 0.0030 | 0.0047 | 0.0046 | 0.0107 | 0.0091 | 0.0039 |
| mnist-1-5 | 0.0000 | 0.0058 | 0.0027 | 0.0027 | 0.0035 | 0.0056 | 0.0068 | 0.0032 | 0.0053 | 0.0041 | 0.0077 |
| mnist-2-6 | 0.0000 | 0.0009 | 0.0013 | 0.0052 | 0.0062 | 0.0007 | 0.0043 | 0.0053 | 0.0053 | 0.0071 | 0.0052 |
| mnist | 0.0000 | 0.0050 | 0.0047 | 0.0070 | 0.0039 | 0.0023 | 0.0044 | 0.0068 | 0.0047 | 0.0050 | 0.0021 |
| F-mnist2v5 | 0.0032 | 0.0030 | 0.0036 | 0.0032 | 0.0030 | 0.0042 | 0.0052 | 0.0081 | 0.0054 | 0.0051 | 0.0043 |
| F-mnist3v4 | 0.0015 | 0.0032 | 0.0032 | 0.0055 | 0.0076 | 0.0021 | 0.0066 | 0.0036 | 0.0045 | 0.0051 | 0.0056 |
| F-mnist7v9 | 0.0030 | 0.0026 | 0.0058 | 0.0022 | 0.0018 | 0.0043 | 0.0038 | 0.0029 | 0.0037 | 0.0048 | 0.0039 |
| F-mnist | 0.0018 | 0.0021 | 0.0026 | 0.0041 | 0.0059 | 0.0024 | 0.0036 | 0.0087 | 0.0061 | 0.0033 | 0.0049 |
| cifar10:0v5 | 0.0066 | 0.0058 | 0.0038 | 0.0057 | 0.0059 | 0.0038 | 0.0086 | 0.0021 | 0.0032 | 0.0102 | 0.0051 |
| cifar10:0v6 | 0.0042 | 0.0016 | 0.0010 | 0.0013 | 0.0012 | 0.0018 | 0.0021 | 0.0032 | 0.0038 | 0.0083 | 0.0085 |
| cifar10:4v8 | 0.0033 | 0.0044 | 0.0056 | 0.0086 | 0.0042 | 0.0031 | 0.0054 | 0.0036 | 0.0025 | 0.0056 | 0.0047 |
| **AVERAGE** | 0.0016 | 0.0027 | 0.0033 | 0.0031 | 0.0039 | 0.0020 | 0.0072 | 0.0058 | 0.0049 | 0.0052 | 0.0049 |

Table 15: Standard deviations of max regret for ensemble forests methods.

| dataset | CART | RIGDT-h | GROOT | FPRDT | CoEvoRDT | CoEvoRDT + FPRDT | N CoEvoRDT | ICoEvoRDT | ICoEvoRDT + FPRDT |
|---|---|---|---|---|---|---|---|---|---|
| ionos | 0.0000 | 0.0151 | 0.0204 | 0.0109 | 0.0172 | 0.0172 | 0.0127 | 0.0104 | 0.0102 |
| breast | 0.0000 | 0.0114 | 0.0150 | 0.0088 | 0.0131 | 0.0131 | 0.0099 | 0.0080 | 0.0080 |
| diabetes | 0.0000 | 0.0144 | 0.0179 | 0.0150 | 0.0173 | 0.0172 | 0.0131 | 0.0116 | 0.0114 |
| bank | 0.0000 | 0.0115 | 0.0097 | 0.0111 | 0.0118 | 0.0118 | 0.0087 | 0.0068 | 0.0067 |
| Japan3v4 | 0.0000 | 0.0108 | 0.0131 | 0.0116 | 0.0112 | 0.0110 | 0.0092 | 0.0070 | 0.0069 |
| spam | 0.0000 | 0.0097 | 0.0083 | 0.0099 | 0.0095 | 0.0094 | 0.0084 | 0.0065 | 0.0064 |
| GesDvP | 0.0000 | 0.0104 | 0.0134 | 0.0122 | 0.0133 | 0.0133 | 0.0102 | 0.0074 | 0.0073 |
| har1v2 | 0.0000 | 0.0115 | 0.0112 | 0.0147 | 0.0124 | 0.0123 | 0.0102 | 0.0089 | 0.0088 |
| wine | 0.0000 | 0.0112 | 0.0080 | 0.0138 | 0.0086 | 0.0085 | 0.0075 | 0.0065 | 0.0064 |
| collision-det | 0.0000 | 0.0107 | 0.0109 | 0.0093 | 0.0095 | 0.0093 | 0.0082 | 0.0071 | 0.0070 |
| mnist-1-5 | 0.0000 | 0.0171 | 0.0128 | 0.0183 | 0.0123 | 0.0123 | 0.0088 | 0.0079 | 0.0077 |
| mnist-2-6 | 0.0000 | 0.0077 | 0.0073 | 0.0092 | 0.0091 | 0.0089 | 0.0066 | 0.0049 | 0.0048 |
| mnist | 0.0000 | 0.0091 | 0.0087 | 0.0098 | 0.0090 | 0.0088 | 0.0067 | 0.0055 | 0.0054 |
| F-mnist2v5 | 0.0000 | 0.0126 | 0.0120 | 0.0113 | 0.0112 | 0.0112 | 0.0085 | 0.0061 | 0.0060 |
| F-mnist3v4 | 0.0000 | 0.0114 | 0.0133 | 0.0150 | 0.0114 | 0.0112 | 0.0100 | 0.0078 | 0.0078 |
| F-mnist7v9 | 0.0000 | 0.0189 | 0.0157 | 0.0149 | 0.0172 | 0.0170 | 0.0133 | 0.0101 | 0.0101 |
| F-mnist | 0.0000 | 0.0144 | 0.0133 | 0.0128 | 0.0125 | 0.0125 | 0.0106 | 0.0083 | 0.0083 |
| cifar10:0v5 | 0.0000 | 0.0192 | 0.0207 | 0.0175 | 0.0178 | 0.0177 | 0.0159 | 0.0141 | 0.0140 |
| cifar10:0v6 | 0.0000 | 0.0076 | 0.0074 | 0.0093 | 0.0089 | 0.0087 | 0.0067 | 0.0057 | 0.0057 |
| cifar10:4v8 | 0.0000 | 0.0121 | 0.0191 | 0.0125 | 0.0170 | 0.0169 | 0.0129 | 0.0113 | 0.0111 |
| AVERAGE | 0.0000 | 0.0123 | 0.0129 | 0.0124 | 0.0125 | 0.0124 | 0.0099 | 0.0081 | 0.0080 |

Table 16: Standard deviations of adversarial accuracy for single decision trees. The best results are bolded.

| dataset | CART | RIGDT-h | GROOT | FPRDT | CoEvoRDT | CoEvoRDT + FPRDT | N CoEvoRDT | ICoEvoRDT | ICoEvoRDT + FPRDT |
|---|---|---|---|---|---|---|---|---|---|
| ionos | 0.0000 | 0.0090 | 0.0097 | 0.0037 | 0.0117 | 0.0104 | 0.0066 | 0.0043 | 0.0039 |
| breast | 0.0000 | 0.0036 | 0.0067 | 0.0069 | 0.0083 | 0.0060 | 0.0066 | 0.0042 | 0.0047 |
| diabetes | 0.0000 | 0.0060 | 0.0068 | 0.0113 | 0.0068 | 0.0115 | 0.0053 | 0.0091 | 0.0086 |
| bank | 0.0000 | 0.0092 | 0.0060 | 0.0088 | 0.0070 | 0.0070 | 0.0049 | 0.0051 | 0.0040 |
| Japan3v4 | 0.0000 | 0.0080 | 0.0083 | 0.0069 | 0.0054 | 0.0059 | 0.0048 | 0.0034 | 0.0023 |
| spam | 0.0000 | 0.0057 | 0.0039 | 0.0060 | 0.0073 | 0.0054 | 0.0026 | 0.0039 | 0.0029 |
| GesDvP | 0.0000 | 0.0065 | 0.0056 | 0.0079 | 0.0106 | 0.0096 | 0.0057 | 0.0049 | 0.0027 |
| har1v2 | 0.0000 | 0.0067 | 0.0059 | 0.0055 | 0.0049 | 0.0093 | 0.0054 | 0.0062 | 0.0064 |
| wine | 0.0000 | 0.0060 | 0.0040 | 0.0065 | 0.0042 | 0.0035 | 0.0053 | 0.0029 | 0.0045 |
| collision-det | 0.0000 | 0.0039 | 0.0039 | 0.0073 | 0.0031 | 0.0052 | 0.0025 | 0.0046 | 0.0045 |
| mnist-1-5 | 0.0000 | 0.0124 | 0.0061 | 0.0119 | 0.0065 | 0.0078 | 0.0040 | 0.0033 | 0.0047 |
| mnist-2-6 | 0.0000 | 0.0034 | 0.0033 | 0.0047 | 0.0041 | 0.0040 | 0.0039 | 0.0031 | 0.0034 |
| mnist | 0.0000 | 0.0040 | 0.0055 | 0.0047 | 0.0059 | 0.0046 | 0.0047 | 0.0029 | 0.0041 |
| F-mnist2v5 | 0.0000 | 0.0086 | 0.0060 | 0.0085 | 0.0064 | 0.0081 | 0.0051 | 0.0036 | 0.0019 |
| F-mnist3v4 | 0.0000 | 0.0065 | 0.0096 | 0.0061 | 0.0066 | 0.0088 | 0.0039 | 0.0058 | 0.0061 |
| F-mnist7v9 | 0.0000 | 0.0116 | 0.0116 | 0.0114 | 0.0055 | 0.0125 | 0.0080 | 0.0032 | 0.0054 |
| F-mnist | 0.0000 | 0.0066 | 0.0069 | 0.0067 | 0.0048 | 0.0046 | 0.0067 | 0.0038 | 0.0050 |
| cifar10:0v5 | 0.0000 | 0.0076 | 0.0129 | 0.0102 | 0.0091 | 0.0102 | 0.0080 | 0.0079 | 0.0094 |
| cifar10:0v6 | 0.0000 | 0.0027 | 0.0039 | 0.0053 | 0.0038 | 0.0026 | 0.0048 | 0.0038 | 0.0029 |
| cifar10:4v8 | 0.0000 | 0.0047 | 0.0087 | 0.0073 | 0.0124 | 0.0107 | 0.0076 | 0.0081 | 0.0087 |
| AVERAGE | 0.0000 | 0.0066 | 0.0068 | 0.0074 | 0.0067 | 0.0074 | 0.0053 | 0.0047 | 0.0048 |

Table 17: Standard deviations of max regret for single decision trees. The best results are bolded.



\clearpage

\bibliography{aaai}

\begin{thebibliography}{26}
\providecommand{\natexlab}[1]{#1}

\bibitem[{Andriushchenko and Hein(2019)}]{andriushchenko2019provably}
Andriushchenko, M.; and Hein, M. 2019.
\newblock Provably robust boosted decision stumps and trees against adversarial attacks.
\newblock \emph{Advances in neural information processing systems}, 32.

\bibitem[{Banfield et~al.(2006)Banfield, Hall, Bowyer, and Kegelmeyer}]{banfield2006comparison}
Banfield, R.~E.; Hall, L.~O.; Bowyer, K.~W.; and Kegelmeyer, W.~P. 2006.
\newblock A comparison of decision tree ensemble creation techniques.
\newblock \emph{IEEE transactions on pattern analysis and machine intelligence}, 29(1): 173--180.

\bibitem[{Breiman(2001)}]{breiman2001random}
Breiman, L. 2001.
\newblock Random forests.
\newblock \emph{Machine learning}, 45: 5--32.

\bibitem[{Breiman(2017)}]{breiman2017classification}
Breiman, L. 2017.
\newblock \emph{Classification and regression trees}.
\newblock Routledge.

\bibitem[{Bull et~al.(2007)Bull, Studley, Bagnall, and Whittley}]{bull2007learning}
Bull, L.; Studley, M.; Bagnall, A.; and Whittley, I. 2007.
\newblock Learning classifier system ensembles with rule-sharing.
\newblock \emph{IEEE transactions on evolutionary computation}, 11(4): 496--502.

\bibitem[{Chen et~al.(2019)Chen, Zhang, Boning, and Hsieh}]{chen2019robust}
Chen, H.; Zhang, H.; Boning, D.; and Hsieh, C.-J. 2019.
\newblock Robust decision trees against adversarial examples.
\newblock In \emph{International Conference on Machine Learning}, 1122--1131. PMLR.

\bibitem[{Codenotti, De~Rossi, and Pagan(2008)}]{codenotti2008experimental}
Codenotti, B.; De~Rossi, S.; and Pagan, M. 2008.
\newblock An experimental analysis of lemke-howson algorithm.
\newblock \emph{arXiv preprint arXiv:0811.3247}.

\bibitem[{Freund, Schapire et~al.(1996)}]{freund1996experiments}
Freund, Y.; Schapire, R.~E.; et~al. 1996.
\newblock Experiments with a new boosting algorithm.
\newblock In \emph{icml}, volume~96, 148--156. Citeseer.

\bibitem[{Fudenberg and Kreps(1993)}]{fudenberg1993learning}
Fudenberg, D.; and Kreps, D.~M. 1993.
\newblock Learning mixed equilibria.
\newblock \emph{Games and economic behavior}, 5(3): 320--367.

\bibitem[{Gong et~al.(2015)Gong, Chen, Zhan, Zhang, Li, Zhang, and Li}]{gong2015distributed}
Gong, Y.-J.; Chen, W.-N.; Zhan, Z.-H.; Zhang, J.; Li, Y.; Zhang, Q.; and Li, J.-J. 2015.
\newblock Distributed evolutionary algorithms and their models: A survey of the state-of-the-art.
\newblock \emph{Applied Soft Computing}, 34: 286--300.

\bibitem[{Guo et~al.(2022)Guo, Teng, Gao, and Zhou}]{guo2022fast}
Guo, J.-Q.; Teng, M.-Z.; Gao, W.; and Zhou, Z.-H. 2022.
\newblock Fast Provably Robust Decision Trees and Boosting.
\newblock In \emph{International Conference on Machine Learning}, 8127--8144. PMLR.

\bibitem[{Kantchelian, Tygar, and Joseph(2016)}]{kantchelian2016evasion}
Kantchelian, A.; Tygar, J.~D.; and Joseph, A. 2016.
\newblock Evasion and hardening of tree ensemble classifiers.
\newblock In \emph{International Conference on Machine Learning}, 2387--2396. PMLR.

\bibitem[{Knight and Campbell(2018)}]{nashpy}
Knight, V.; and Campbell, J. 2018.
\newblock Nashpy: A Python library for the computation of Nash equilibria.
\newblock \emph{Journal of Open Source Software}, 3(30): 904.

\bibitem[{Lemke and Howson(1964)}]{lemke1964equilibrium}
Lemke, C.~E.; and Howson, J.~T., Jr. 1964.
\newblock Equilibrium points of bimatrix games.
\newblock \emph{Journal of the Society for Industrial and Applied Mathematics}, 12(2): 413--423.

\bibitem[{Luong, Melab, and Talbi(2010)}]{luong2010gpu}
Luong, T.~V.; Melab, N.; and Talbi, E.-G. 2010.
\newblock GPU-based island model for evolutionary algorithms.
\newblock In \emph{Proceedings of the 12th annual conference on Genetic and evolutionary computation}, 1089--1096.

\bibitem[{Michalewicz(2013)}]{michalewicz2013genetic}
Michalewicz, Z. 2013.
\newblock \emph{Genetic algorithms + data structures = evolution programs}.
\newblock Springer Science \& Business Media.

\bibitem[{Ranzato and Zanella(2021)}]{ranzato2021genetic}
Ranzato, F.; and Zanella, M. 2021.
\newblock Genetic adversarial training of decision trees.
\newblock In \emph{Proceedings of the Genetic and Evolutionary Computation Conference}, 358--367.

\bibitem[{Rokach and Maimon(2005)}]{rokach2005decision}
Rokach, L.; and Maimon, O. 2005.
\newblock Decision trees.
\newblock \emph{Data mining and knowledge discovery handbook}, 165--192.

\bibitem[{Savage(1951)}]{savage1951theory}
Savage, L.~J. 1951.
\newblock The theory of statistical decision.
\newblock \emph{Journal of the American Statistical association}, 46(253): 55--67.

\bibitem[{Skolicki(2005)}]{skolicki2005analysis}
Skolicki, Z. 2005.
\newblock An analysis of island models in evolutionary computation.
\newblock In \emph{Proceedings of the 7th annual workshop on Genetic and evolutionary computation}, 386--389.

\bibitem[{Skolicki and De~Jong(2004)}]{skolicki2004improving}
Skolicki, Z.; and De~Jong, K. 2004.
\newblock Improving evolutionary algorithms with multi-representation island models.
\newblock In \emph{International conference on parallel problem solving from nature}, 420--429. Springer.

\bibitem[{Tanese(1989)}]{tanese1989distributed}
Tanese, R. 1989.
\newblock \emph{Distributed genetic algorithms for function optimization}.
\newblock University of Michigan.

\bibitem[{v.~Neumann(1928)}]{v1928theorie}
v.~Neumann, J. 1928.
\newblock Zur theorie der gesellschaftsspiele.
\newblock \emph{Mathematische annalen}, 100(1): 295--320.

\bibitem[{Vos and Verwer(2021)}]{vos2021efficient}
Vos, D.; and Verwer, S. 2021.
\newblock Efficient training of robust decision trees against adversarial examples.
\newblock In \emph{International Conference on Machine Learning}, 10586--10595. PMLR.

\bibitem[{Yu and Gen(2010)}]{yu2010introduction}
Yu, X.; and Gen, M. 2010.
\newblock \emph{Introduction to evolutionary algorithms}.
\newblock Springer Science \& Business Media.

\bibitem[{Żychowski, Perrault, and Mańdziuk(2024)}]{zychowski2024CoEvoRDT}
Żychowski, A.; Perrault, A.; and Mańdziuk, J. 2024.
\newblock Coevolutionary Algorithm for Building Robust Decision Trees under Minimax Regret.
\newblock \emph{Proceedings of the AAAI Conference on Artificial Intelligence}, 38(19): 21869--21877.

\end{thebibliography}

\end{document}